\documentclass[10pt,journal,compsoc]{IEEEtran}

\usepackage{amssymb}
\usepackage{xcolor}
\usepackage{booktabs}
\usepackage{multirow}
\usepackage{hyperref}
\usepackage{booktabs, multirow}
\usepackage{soul}
\usepackage{xcolor,colortbl}
\usepackage[numbers]{natbib}
\usepackage{tikz}
\usepackage{subcaption}

\usepackage{amsmath}

\begin{document}




\title{A Survey on Semi-Supervised Semantic Segmentation}






\author{Adrián~Peláez-Vegas,~\IEEEmembership{}
        Pablo~Mesejo,~\IEEEmembership{}
        and~Julián~Luengo~\IEEEmembership{}
\IEEEcompsocitemizethanks{\IEEEcompsocthanksitem Adrián Peláez-Vegas, Pablo Mesejo and Julián Luengo are with the Department of Computer Science and Artificial Intelligence, Andalusian Research Institute in Data Science and Computational Intelligence, DaSCI, University of Granada, 18071, Granada, Spain. 

E-mail: adrianpelaez@ugr.es, \{pmesejo, julianlm\}@decsai.ugr.es  }
\thanks{}}

\markboth{Journal of \_ Files,~Vol.~x, No.~x, Month~Year}%
{Shell \MakeLowercase{\textit{et al.}}: Bare Advanced Demo of IEEEtran.cls for IEEE Computer Society Journals}

\IEEEtitleabstractindextext{%

\begin{abstract}
Semantic segmentation is one of the most challenging tasks in computer vision. However, in many applications, a frequent obstacle is the lack of labeled images, due to the high cost of pixel-level labeling. In this scenario, it makes sense to approach the problem from a semi-supervised point of view, where both labeled and unlabeled images are exploited. In recent years this line of research has gained much interest and many approaches have been published in this direction. Therefore, the main objective of this study is to provide an overview of the current state of the art in semi-supervised semantic segmentation, offering an updated taxonomy of all existing methods to date. This is complemented by an experimentation with a variety of models representing all the categories of the taxonomy on the most widely used becnhmark datasets in the literature, and a final discussion on the results obtained, the challenges and the most promising lines of future research.
\end{abstract}

}

\maketitle

\IEEEdisplaynontitleabstractindextext

\IEEEpeerreviewmaketitle



\section{Introduction}
\label{sec:introduction}

\begin{figure*}\centering
\begin{subfigure}[h]{0.24\textwidth}
    \includegraphics[width=.99\textwidth]{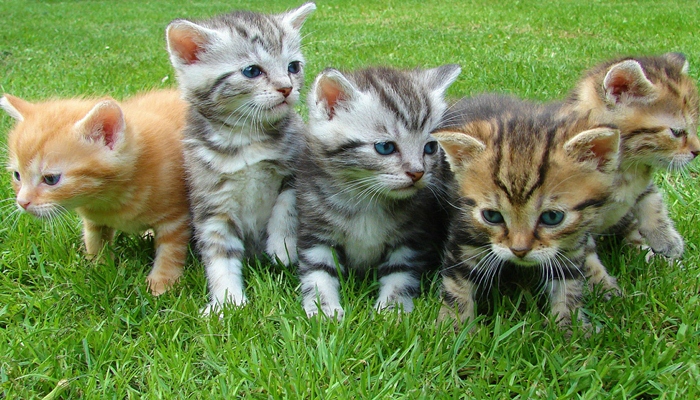}\caption{\scriptsize Original image}
\end{subfigure}
\begin{subfigure}[h]{0.24\textwidth}
    \includegraphics[width=.99\textwidth]{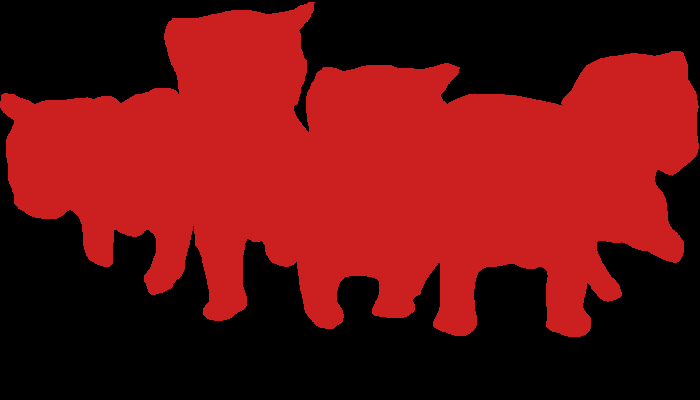}\caption{\scriptsize Semantic segmentation}
\end{subfigure}
\begin{subfigure}[h]{0.24\textwidth}
    \includegraphics[width=.99\textwidth]{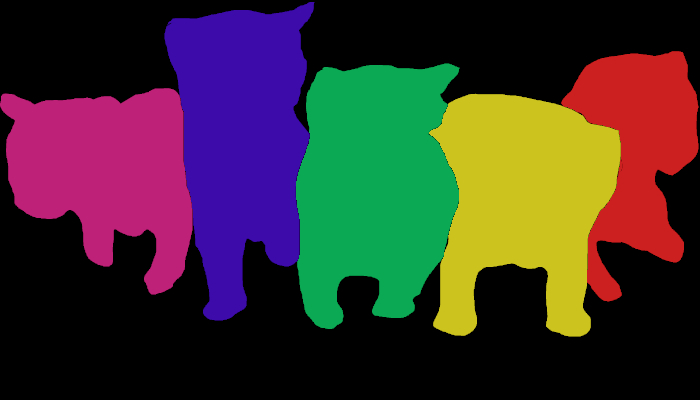}\caption{\scriptsize Instance segmentation}
\end{subfigure}
    \caption{ \small Visual representation of the different variants of the image segmentation problem.}\label{fig: computer_vision_problems}
\end{figure*}

\IEEEPARstart{I}{mage} Segmentation is one of the oldest and most widely studied computer vision (CV) problems \cite{szeliski2010computer, Minaee2022ImageSU}. It consists of dividing an image into different non-overlapping regions and assigning the corresponding label to each pixel in the image. This task can be considered as a pixel-level classification problem, which leads to a significant increase in complexity compared to other CV problems, such as image-level classification or object detection \cite{szeliski2010computer}. We can differentiate between two different types of image segmentation problems. On the one hand, semantic segmentation classifies each pixel with the corresponding semantic class, thus giving the same class label to all objects or regions of the image that belong to this class. On the other hand, instance segmentation attempts to go one step further and tries to distinguish between different occurrences of the same class (see Figure \ref{fig: computer_vision_problems}). This paper focuses on semantic segmentation (SS), which has gained much interest in recent years with important applications in different areas such as medical imaging \cite{medley2021cycoseg}, autonomous driving \cite{Orsic2019in}, aerial scene analysis \cite{Mou2019a} or metallographic images \cite{Luengo2021a}, among others \cite{katircioglu2021self, sakaridis2020map}.

\begin{figure}[]
  \centering
  \includegraphics[width=0.5\textwidth]{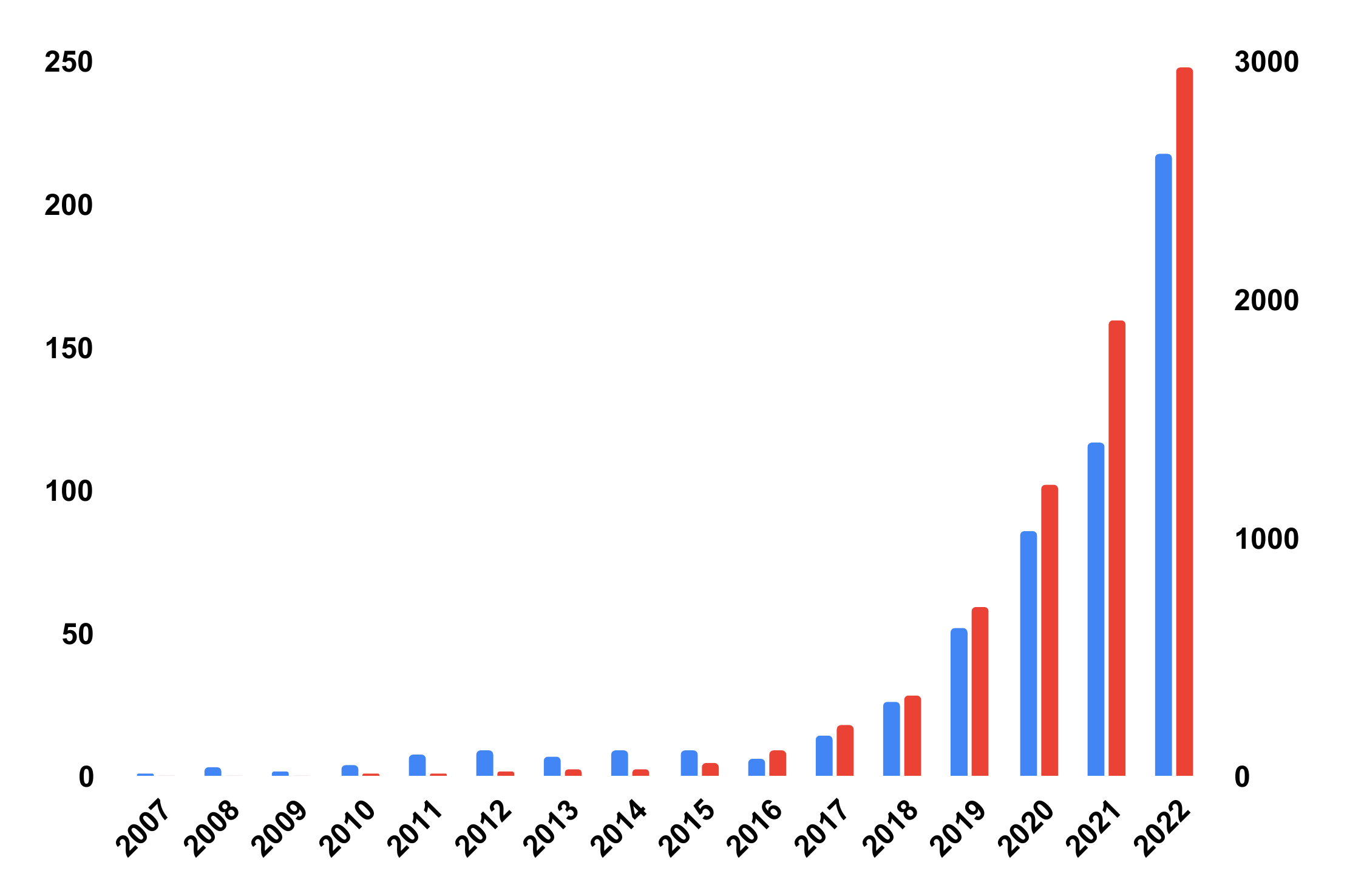}
  \caption{ \small Histogram of publications (blue) and citations (red) related to semi-supervised SS. The left axis represents the number of publications, and the right axis the number of citations. The queries, run in Elsevier Scopus on the 18th of January, 2023, showing a clear growing tendency.}\label{fig: histogram}
\end{figure}

Methods based on deep learning (DL) have recently shown great potential \cite{lecun2015deep, Goodfellow2016Deep, Schmidhuber2015DeepLI}, becoming the state-of-the-art methods in many CV problems \cite{Liu2019DeepLF, Minaee2022ImageSU, Hesamian2019DeepLT, Wang2021DeepLF, Wang2021DeepFR}. The SS problem has traditionally been addressed by classical image processing and CV techniques, such as thresholding techniques or clustering algorithms \cite{Sezgin2004SurveyOI, Chen1998ImageSV}. However, with the emergence of DL methods it has been possible to give a leap of quality in the segmentation results, as in many other CV problems. That is why all the semi-supervised methods that form the state of the art in semi-supervised SS, and that we include in this study, are based on DL.

Depending on the degree of detail of the ground truth labeling (i.e. the output considered correct) and the portion of labeled examples in relation to the total number of available images we can face different scenarios: fully supervised scenario, weakly supervised scenario, semi-supervised scenario and unsupervised scenario. Due to the difficulty and effort involved in labeling images at the pixel level, approaches based on semi-supervised learning (SSL) \cite{Chapelle2006SemiSupervisedL}, in which we have a reduced amount of labeled images, and a larger amount of unlabeled images, are becoming more and more relevant. These semi-supervised methods extract knowledge from the labeled data in a supervised way, and in an unsupervised way from the unlabeled data, thus reducing the labeling effort required in a fully supervised scenario, and obtaining notably better results than in an unsupervised scenario.

To the best of our knowledge, there is only one survey that tries to address  semi-supervised SS methods \cite{a2020Zhang}. However, the aforementioned review does not cover certain aspects that we consider important, and that we address in our work:
\begin{itemize}
    \item First of all, this previous study, published in 2019, does not include the methods proposed in recent years, when the problem has gained more interest. Figure \ref{fig: histogram} shows a histogram where it is clear that the majority of publications and citations in this field are concentrated in the last two years, which are outside the scope of the previous survey.
    \item Secondly, it does not focus exclusively on the semi-supervised scenario, but also includes the methods proposed for a weakly supervised scenario. It does not go into sufficient depth for the semi-supervised field, which is currently a sufficiently broad scenario to be addressed exclusively.
    \item Finally, it does not include an experimental study that allows a fair comparison between methods and allows the reader to have a clear idea about the performance of each one of them.
\end{itemize}
Other surveys related to our target field, but not totally focused on it, have been proposed recently. Several of these works focus on the SSL paradigm. Some of them review methods based on DL \cite{Yang2021ASO, Ouali2020AnOO}, and others carry out the review of methods from a general point of view \cite{Triguero2013SelflabeledTF, vanEngelen2019ASO}. These studies take the image classification problem as the basic problem, so they do not cover the wide field of semi-supervised SS that we try to tackle in this study. On the other hand, there are other surveys that focus on the SS problem, but do not address the semi-supervised scenario or treat it very superficially \cite{Minaee2022ImageSU, GarciaGarcia2018ASO, Hao2020ABS}.

The main contributions of this work are summarized as follows:
\begin{itemize}
    \item We provide an updated taxonomy of semi-supervised SS methods as well as a description of them.
    \item We carry out an experimentation with a wide range of state-of-the-art semi-supervised segmentation methods on the most widely used datasets in the literature.
    \item A discussion is proposed on the results obtained, advantages and shortcomings of the current methods, challenges and future lines of work in this field.
\end{itemize}

The content of this article is organized as follows. The key concepts and background about the problem under discussion, as well as the existing datasets, are presented in section \ref{sec:background}. Then, in section \ref{sec:taxonomy}, the proposed taxonomy is described, followed by a subsection for each of the categories that form our taxonomy, where we go into detail about the different method proposals belonging to each category. A detailed description of the proposed experimentation as well as the results obtained are shown and discussed in section \ref{sec:results}, while a reflection on the main difficulties, challenges and future directions is presented in section \ref{sec:challenges}. Finally, section \ref{sec:conclusion} ends with the conclusions obtained.

\section{Background}
\label{sec:background}

\subsection{Problem formulation}  The SSL paradigm is halfway between fully supervised learning and unsupervised learning, and it deals with data sets that are just partly annotated. Moreover, the ratio between the amount of labeled and unlabeled data is usually, in real-world problems, very unfavorable for the labeled part. Specifically, in semantic image segmentation, the imbalancement between labeled and unlabeled data is often even more frequent and pronounced due to the difficulty of annotating an image at the pixel level \cite{Chapelle2006SemiSupervisedL}.

In this context, we have a dataset $X = \{X_L, X_U\}$ where $X_L = \{ (x_i, y_i) \}_{i=1}^l $ is the subset of labeled data and $X_U = \{ x_i \}_{i=1}^u$ is the subset of unlabeled data, $x_i$ is an input image and $y_i$ its corresponding label map, $l$ and $u$ are the number of labeled and unlabeled data, respectively, and commonly $l << u$. The goal of a semi-supervised approach is to extract knowledge from both, the labeled and unlabeled data, in order to obtain a model with better performance than the one we could get from training only with labeled data.

\subsection{Classical approaches to SS}

In this section we introduce those methods for SS that were proposed before the DL era (i.e., before 2012). Although our study focuses on deep segmentation models, it is relevant to analyze prior art. We consider three levels of supervision: fully-supervised, unsupervised, and semi-supervised.

It is important to note that the first methods proposed for image segmentation are essentially unsupervised. Among these methods we can find basic image processing and CV techniques that stand out for their simplicity and efficiency when applied. Among them we find methods such as image thresholding \cite{Sezgin2004SurveyOI}, region growing \cite{Trmeau1997ARG} and deformable models \cite{Terzopoulos2005DeformableM}. On the other hand, the application of unsupervised machine learning algorithms, such as clustering algorithms (e.g. k-means) \cite{Chen1998ImageSV} or graph-based models \cite{Felzenszwalb2004EfficientGI}, has also been proposed.

Subsequently, proposals based on supervised machine learning algorithms started to emerge. For instance, Random Forest \cite{Schroff2008ObjectCS}, SVMs \cite{Felzenszwalb2009ObjectDW} and conditional or Markov random fields \cite{Moser2012MarkovRF, Gonfaus2010HarmonyPF} were proposed and adapted to the image segmentation problem. 


The semi-supervised setting was the last to be addressed. Some extensions of fully supervised methods were proposed to equip them with the ability to handle unlabeled data. To the best of our knowledge, the first method proposed specifically for semi-supervised segmentation was a mixed model based on a tree-structured patch-based approach and the random forest algorithm \cite{Badrinarayanan2013SemiSupervisedVS}. A model based on weighted graphs was also proposed for the semi-supervised segmentation of 3D surfaces \cite{Bergamasco2012AGT}. Due to its fast semi-supervised classification and its interpretability, the random forest algorithm is used in other works to address the semi-supervised segmentation problem, as is the case of \cite{Mahapatra2013SemiSupervisedAA} where the authors propose its use on abdominal magnetic resonance. Finally, a last proposal prior to the emergence of DL proposes a method that incorporates Gaussian mixture models, random walk models and SVMs \cite{Tian2014SemisupervisedLF}.

\subsection{Deep learning for SS} 


The performance of the semi-supervised methods largely depend on the good choice, fit and training of the supervised model on which it is based. Therefore, in this section we present the background of these supervised segmentation models.

Initially, DL techniques, generally convolutional neural networks (CNN)  \cite{rawat2017deep}, were proposed and applied to the problem of image classification, obtaining a leap in quality with respect to the traditional techniques that had been used until then. Due to the good results obtained, these techniques were extended to other areas of CV, trying to solve increasingly complex and fine-grained problems, such as object detection \cite{Redmon2016YouOL} and segmentation \cite{Ronneberger2015UNetCN, Chen2017RethinkingAC}.

The key idea underlying most DL models for SS is the fully convolutional neural networks (FCNN), proposed in \cite{Shelhamer2015FullyCN}. In this work, the approach proposed by the authors consists in reusing well-known CNN (such as VGG \cite{Simonyan2015VeryDC}, ResNet \cite{He2016DeepRL} or EfficientNet \cite{Tan2019EfficientNetRM}) originally proposed and applied in image classification problems, adapting them to address the SS problem. This adaptation, in general terms, consists in replacing the final fully connected layers of these models by convolutional layers, thus obtaining as output feature maps instead of a vector of classification scores. Finally, the resulting feature maps are upsampled by using deconvolution operations \cite{Zeiler2011AdaptiveDN} to obtain the final segmentation map. FCNN achieves a performance gains (20\% improvement) in Pascal VOC \cite{Everingham2009ThePV}, one of the main SS benchmarks. 

The FCNN approach demonstrated that the problem of SS could be addressed through DL techniques, thus opening a new line of research that today is in a really advanced state, with many new methods that improve the original proposal of FCNN. The main difference between these methods lies in the way they upsample the output of the convolutional network to obtain the final segmentation map. For instance, encoder-decoder architectures (e.g. U-Net model \cite{Ronneberger2015UNetCN}) chain a decoder to the CNN. Another well-known example is the DeepLab model \cite{Chen2018EncoderDecoderWA, Chen2018DeepLabSI, Chen2017RethinkingAC} that uses atrous convolution to increase its range of vision and increase its capacity to capture contextual information.


    
    
    
    

\subsection{Datasets} 
\label{subsec:datasets}


The availability of pixel-level annotated datasets to address the SS problem is not as high as the annotations needed for other CV problems, such as image classification that only needs image-level labeling, mainly due to the difficulty of performing such annotations manually.

For instance, to annotate a single image from the Cityscapes dataset requires approximately three hours of manual work \cite{Cordts2016TheCD}. However, due to the interest that this problem has raised in recent years, efforts have been made to annotate datasets to train segmentation models, and we currently have a wide range of these datasets of different types and domains.


In this sense, this subsection exposes the datasets available for SS and classifies them by images content. A more detailed description of the datasets used in our experimentation is also provided. It should be noted that each of these datasets can be used as a benchmark in a fully supervised scenario, as well as in a semi-supervised or unsupervised scenario, just by choosing for which data the corresponding labels are used in the training process. Since this review focuses on the semi-supervised setting, we also expose which of the existing datasets have been used as semi-supervised benchmark, as well as the most frequently labeled and unlabeled data partitions used in the literature. Table \ref{tab: datasets} displays commonly employed datasets in SS.

\begin{table}[!htp]\centering
\caption{ \small Summary table of the most widely used datasets for SS classified according to the nature of the images.}\label{tab: datasets}
\scriptsize
\bgroup
\def\arraystretch{1.2}
\begin{tabular}{|c|c|}\hline
\textbf{Images content} &\textbf{Datasets} \\\hline \hline

\multirow{4}{*}{General images} &PASCAL VOC 2012 \cite{Everingham2009ThePV}, CIFAR-10/100 \cite{Krizhevsky2009LearningML}, \\ 
&SegTrack v2 \cite{Li2013VideoSB}, PASCAL Context \cite{Mottaghi2014TheRO},\\ 
&Microsoft-COCO \cite{Lin2014MicrosoftCC}, ADE20K \cite{Zhou2017ScenePT}, DAVIS \cite{PontTuset2017The2D},\\
&YouTube VIS \cite{Yang2019VideoIS}, YouTube dataset \cite{Prest2012LearningOC}\\
\hline

\multirow{4}{*}{Street Views} &CamVid \cite{Brostow2009SemanticOC}, Cityscapes \cite{Cordts2016TheCD}, \\
&Mapillary Vistas \cite{Neuhold2017TheMV},  KITTI \cite{Geiger2012AreWR}, \\
&Synthia \cite{Ros2016TheSD}, GTA5 \cite{Richter2016PlayingFD}, IDD \cite{Varma2019IDDAD},\\
&Apolloscape-Scene \cite{Huang2018TheAD}\\
\hline

\multirow{3}{*}{Indoor environments} &SUN RGB-D \cite{Song2015SUNRA}, ScanNet \cite{Dai2017ScanNetR3}, \\
& Standford 2D \cite{Armeni2017Joint2D}, NYUD v2 \cite{Silberman2012IndoorSA},\\
& Cornell RGB-D \cite{Koppula2011SemanticLO}\\
\hline

\multirow{3}{*}{Outdoor environments} &INRIA-Graz-02 \cite{Marszalek2007AccurateOL}, Freiburg Forest \cite{Valada2016DeepMS}, \\
&  PASCAL SBD \cite{Hariharan2011SemanticCF}, Sift-Flow \cite{Liu2011NonparametricSP} \\
&LabelMe \cite{Russell2007LabelMeAD}, Microsoft Cambridge \cite{Shotton2011RealtimeHP},\\
\hline

\multirow{3}{*}{Human Pictures} &Adobe's portrait \cite{Shen2016AutomaticPS}, Helen \cite{Le2012InteractiveFF},\\ &LIP \cite{Gong2017LookIP}, DeepFashion 2 \cite{Ge2019DeepFashion2AV},\\
& Open EDS \cite{Garbin2019OpenEDSOE}\\
\hline

Material images &MINC \cite{Bell2015MaterialRI}, UHCS \cite{DeCost2019HighTQ}, MetalDAM \cite{Luengo2022ATO} \\
\hline

 \multirow{2}{*}{\begin{tabular}[c]{@{}c@{}} Satellite and \\ aerial images\end{tabular}} &EuroSAT \cite{Helber2019EuroSATAN}, FloodNet \cite{Rahnemoonfar2021FloodNetAH}, xBD \cite{Gupta2019CreatingXA}, \\ 
&AIRS \cite{Chen2019AerialIF}, GID \cite{Tong2018LandcoverCW}, iSAID \cite{Zamir2019iSAIDAL} \\
\hline

\multirow{3}{*}{Medical images} & Medical Segmentation Decathlon \cite{Simpson2019ALA}\\
&Drive \cite{Staal2004RidgebasedVS}, GlaS \cite{Sirinukunwattana2017GlandSI}, CoNSeP \cite{Graham2019HoverNetSS}, \\
&Kvasir-SEG \cite{Jha2020KvasirSEGAS}, REFUGE \cite{Orlando2020REFUGECA}, \\
&BratS \cite{Menze2015TheMB}, PROMISE 12 \cite{Litjens2014EvaluationOP}, \\ 
\hline

\end{tabular}
\egroup

\end{table}

With the aim of providing comparable results with other studies, the research community mainly uses a small subset of the datasets presented as benchmarks for their proposals. Below we provide a description of the two most frequent datasets and the partitions commonly used in the semi-supervised scenario.

\begin{itemize}
    \item \textbf{PASCAL VOC 2012} \cite{Everingham2009ThePV}: The PASCAL VOC 2012 dataset\footnote{\href{http://host.robots.ox.ac.uk/pascal/VOC/}{http://host.robots.ox.ac.uk/pascal/VOC/}} is the most widely used as benchmark in SS studies. It is composed of general situation and object-centered images with variable size. This dataset has 20 object classes and an additional background class. The official partitions for training, validation and test consist of 1464, 1449 and 1456 images, respectively. However, an augmented version with 9118 extra images from the Segmentation Boundary Dataset (SBD) \cite{Hariharan2011SemanticCF} is often used, bringing the training set to 10582 images with associated pixel-wise labeling. For the semi-supervised scenario, the following ratios of data in the training set are usually selected as labeled partitions: 1/100 (106 images), 1/50 (212 images), 1/20 (529 images), 1/8 (1323 images) and 1/4 (2646 images). For the rest of the images their labels are not be taken into account and form the unsupervised data input in semi-supervised methods.
    
    \item \textbf{Cityscapes} \cite{Cordts2016TheCD}: The Cityscapes dataset\footnote{\href{https://www.cityscapes-dataset.com/}{https://www.cityscapes-dataset.com/}} is another of the most widely used datasets and, specifically, one of the most important ones for autonomous driving applications. This dataset is composed of a series of sequential street view images taken from a vehicle in different European cities, with size 2048 × 1024 and 19 classes. The official partitions for training, validation and test are composed of 2975, 500 and 1525 images respectively. For the semi-supervised scenario, the following ratios of data in the training set are usually selected as labeled partitions: 1/16 (186 images), 1/8 (372 images), 1/4 (744 images) and 1/2 (1488 images).
    
\end{itemize}


\section{Semi-Supervised Semantic Segmentation Methods}
\label{sec:taxonomy}

In this section, a review and explanation of the techniques proposed for the semi-supervised segmentation problem and a taxonomyzation is carried out. First, the proposed taxonomy is presented, followed by a detailed section for each of its categories, with detailed explanations of the representative methods.

\subsection{Taxonomy}

\begin{figure}[h]
  \centering
  \includegraphics[width=0.47\textwidth]{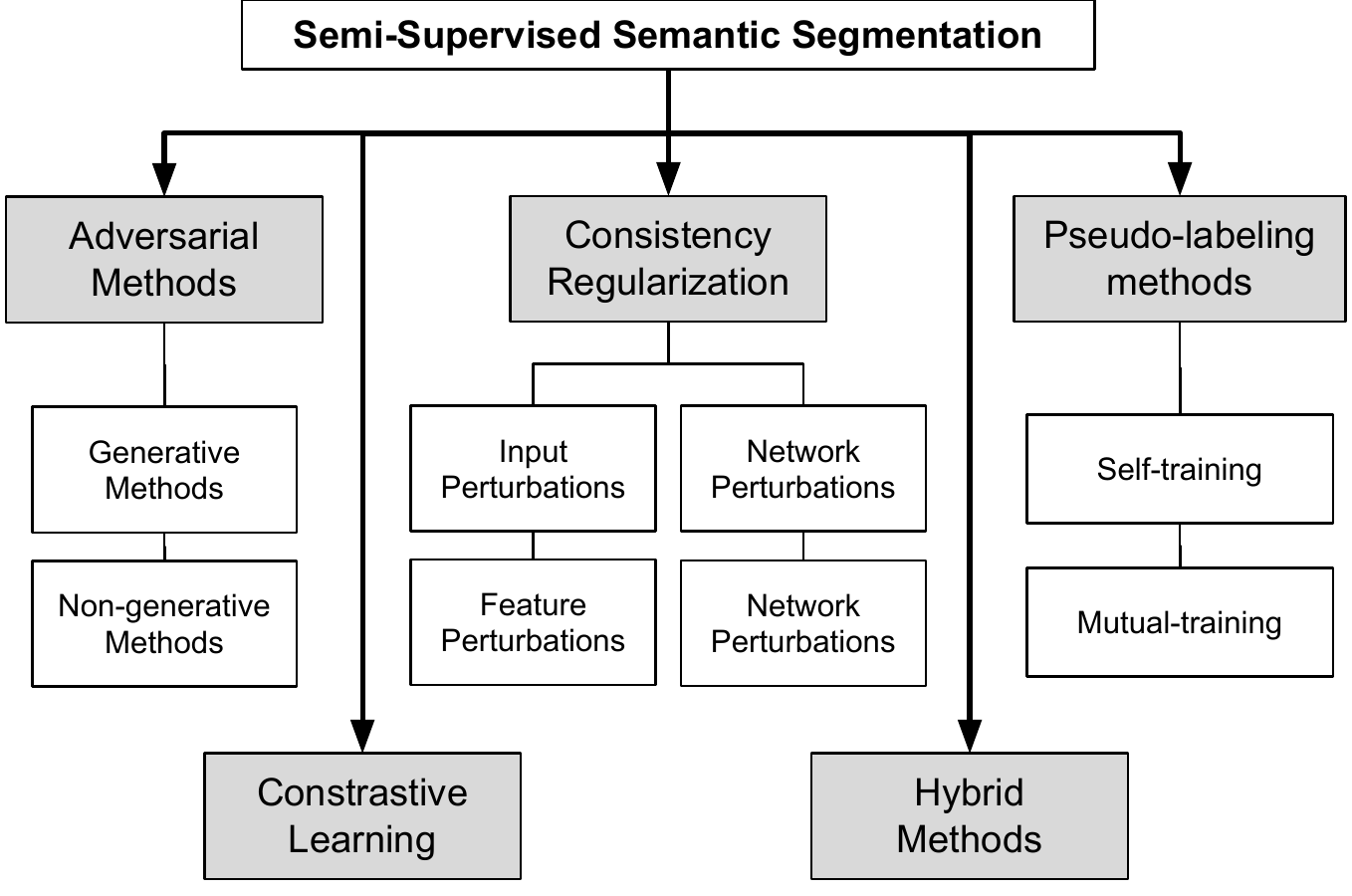}
  \caption{\small Dendrogram showing the taxonomy proposed in this paper. References to the methods belonging to each category are located next to each leaf node.}
  \label{fig: dendogram}
\end{figure}

According to the nature and main characteristics of the existing methods in the semi-supervised SS literature, we propose a taxonomy that classifies these methods into five categories. This taxonomy is represented graphically by the dendogram displayed in Figure \ref{fig: dendogram}, and a list of all existing methods in each category is provided in Table \ref{tab: methods}.

\begin{itemize}

\item  The first category includes those methods that adopt a GAN-like structure and adversarial training between two networks, one acting as a generator and the other as a discriminator (Section \ref{subsec: adversarial}). 


\item The next category corresponds to consistency regularization methods. These methods include a regularization term in the loss function to minimize the differences between different predictions of the same image, which are obtained by applying perturbations to the images or to the models involved (Section \ref{subsec: consistency}).


\item Another category comprises those methods that are based on pseudo-labeling of unlabeled data. In general terms, these methods rely on predictions previously made on the unlabeled data with a model trained on the labeled data to obtain pseudo-labels. In this way they are able to include the unlabeled data in the training process (Section \ref{subsec: pseudo-labeling}).


\item The fourth category includes methods based on contrastive learning. This learning paradigm groups similar elements and separates them from dissimilar elements in a certain representation space, often different from the output space of the models (Section \ref{subsec: contrastive}). 


\item Finally, we group in a fifth category those methods that present characteristic elements of several of the previously exposed categories. We can mainly find hybrid methods between consistency regularization, pseudo-labeling and contrastive learning (Section \ref{subsec: hybrid}).


\end{itemize}

\begin{table}[!htp]\centering
\caption{ \small List of existing semi-supervised SS methods in the literature, classified according to the defined taxonomy.}
\scriptsize
\bgroup
\def\arraystretch{1.2}

\begin{tabular}{|c|c|c|c|}
\hline
\textbf{Method}                                      & \textbf{Category}                                    & \textbf{Subcategory}                          & \textbf{Year} \\ \hline \hline
\cite{Li2021SemanticSW}                     & \multirow{11}{*}{\begin{tabular}[c]{@{}c@{}} Adversarial \\ methods\end{tabular}}       & \multirow{2}{*}{Generative}          & 2021          \\
\cite{Souly2017SemiSS}                      &                                             &                                      & 2017                \\ \cline{3-4}
\cite{Jin2021AdversarialNI}                 &                                             &  \multirow{9}{*}{Non-generative}     & 2021             \\
\cite{xu2021semi}                           &                                             &                                      & 2021          \\
\cite{di2021semi}                           &                                             &                                      & 2021          \\
\cite{Mendel2020SemisupervisedSB}           &                                             &                                      & 2020             \\
(GCT) \cite{Ke2020GuidedCT}                 &                                             &                                      & 2020                \\
\cite{Zhang2020RobustAL}                    &                                             &                                      & 2020             \\
(S4GAN) \cite{Mittal2021SemiSupervisedSS}   &                                             &                                      & 2019                \\
\cite{liu2019semi}                          &                                             &                                      & 2019          \\
\cite{Hung2018AdversarialLF}                &                                             &                                      & 2018                \\\hline \hline

(ComplexMix) \cite{chen2021complexmix}      & \multirow{12}{*}{\begin{tabular}[c]{@{}c@{}}Consistency\\ regularization\end{tabular}} & \multirow{6}{*}{\begin{tabular}[c]{@{}c@{}}Input \\perturbations\end{tabular}} & 2021                \\
\cite{grubivsic2021baseline}                &                                             &                                      & 2021        \\
(ClassMix) \cite{Olsson2021ClassMixSD}      &                                             &                                      & 2021                \\
(CutMix) \cite{French2020SemisupervisedSS}  &                                             &                                      & 2020                \\
\cite{li2020semi}                           &                                             &                                      & 2020        \\
\cite{Kim2020StructuredCL}                  &                                             &                                      & 2020             \\\cline{3-4}
(CCT) \cite{Ouali2020SemiSupervisedSS}            &                                             & \begin{tabular}[c]{@{}c@{}}Feature \\perturbations \end{tabular}               & 2020                \\\cline{3-4}
\cite{an2022deep}                           &                                             & \multirow{3}{*}{\begin{tabular}[c]{@{}c@{}}Network \\perturbations\end{tabular}} & 2022                \\
(CPS) \cite{Chen2021SemiSupervisedSS}       &                                             &                                      & 2020          \\
\cite{peng2020deep}                         &                                             &                                      & 2020          \\\cline{3-4}
\cite{wu2022perturbation}                   &                                             & \multirow{2}{*}{\begin{tabular}[c]{@{}c@{}}Combined \\perturbations\end{tabular}}               & 2022   \\
\cite{Liu2021PerturbedAS}                   &                                             &                                      & 2021             \\\hline \hline

(ST++) \cite{Yang2021STMS}                  & \multirow{9}{*}{Pseudo-labeling}            & \multirow{7}{*}{Self-training}       & 2021                \\
(GIST \& RIST) \cite{Teh2021TheGA}          &                                             &                                      & 2021             \\
\cite{li2021residual}                       &                                             &                                      & 2021             \\
\cite{Yuan2021ASB}                          &                                             &                                      & 2021             \\
\cite{he2021re}                             &                                             &                                      & 2021             \\
\cite{Zhu2020ImprovingSS}                   &                                             &                                      & 2020             \\
\cite{chen2020digging}                      &                                             &                                      & 2020            \\\cline{3-4}

(DMT) \cite{feng2020dmt}                    &                                             & \multirow{2}{*}{Mutual-training}                      & 2022                \\
\cite{zhou2022catastrophic}                 &                                             &                                      & 2022 \\\hline \hline

(ReCo) \cite{Liu2021BootstrappingSS}        & \multirow{2}{*}{\begin{tabular}[c]{@{}c@{}}Contrastive\\ learning\end{tabular} }       & \multirow{2}{*}{-}                   & 2021 \\
\cite{Alonso2021SemiSupervisedSS}           &                                             &                                      & 2021                \\\hline \hline
(CTT) \cite{xiao2022semi}                   & \multirow{9}{*}{Hybrid}                     & \multirow{8}{*}{-}                   & 2022                \\
\cite{cao2022adversarial}                   &                                             &                                      & 2022                 \\
(AEL) \cite{Hu2021SemiSupervisedSS}         &                                             &                                      & 2021                 \\
(GuidedMix-Net) \cite{Tu2021GuidedMixNetLT} &                                             &                                      & 2021                \\
(CAC) \cite{Lai2021SemisupervisedSS}        &                                             &                                      & 2021                \\
\cite{Zhong2021PixelCS}                     &                                             &                                      & 2021             \\ 
\cite{zhou2021c3}                     &                                             &                                      & 2021             \\ 
(PseudoSeg) \cite{Zou2021PseudoSegDP}       &                                             &                                      & 2020                \\
\cite{Ke2022ATS}                            &                                             &                                      & 2020             \\\hline 

\end{tabular}
\egroup
\label{tab: methods}
\end{table}

\subsection{Adversarial methods}\label{subsec: adversarial}

Generative adversarial networks (GANs) \cite{Goodfellow2014GenerativeAN} have become a very popular framework due to the good performance they have demonstrated in a multitude of problems such as image generation \cite{Radford2016UnsupervisedRL}, object detection \cite{Wang2017AFastRCNNHP} or SS \cite{Luc2016SemanticSU}, among many others. A typical GAN framework consists of two networks, generator and discriminator. The purpose of the generator is to learn the distribution of the target data, thus allowing the generation of synthetic images from random noise. The purpose of the discriminator is to distinguish between real images (belonging to the real distribution) and fake images (generated by the generator). The training process of these networks is carried out in an adversarial way. The generator tries to confuse the discriminator, generating images increasingly similar to the target distribution, and the discriminator attempts to increase its ability to distinguish between real and fake images. This adversary training process is formally defined below:

\begin{equation}\label{eq: minmax}
\small
\begin{split}
\min_G \max_D V(D,G) = &\mathbb{E}_{x  \sim X}[log (D(x))] + \\ &\mathbb{E}_{z  \sim p_z(z)} [log(1-D(G(z)))]
\end{split}
\end{equation}

The Equation \ref{eq: minmax} represents the min max game played by the discriminator $D$ and the generator $G$. The purpose of the first term is to maximize the accuracy obtained by $D$, while the second term attempts to increase the quality of the images generated by $G$, from random noise $z$.

\begin{figure}[h]
  \centering
  \includegraphics[width=0.47\textwidth]{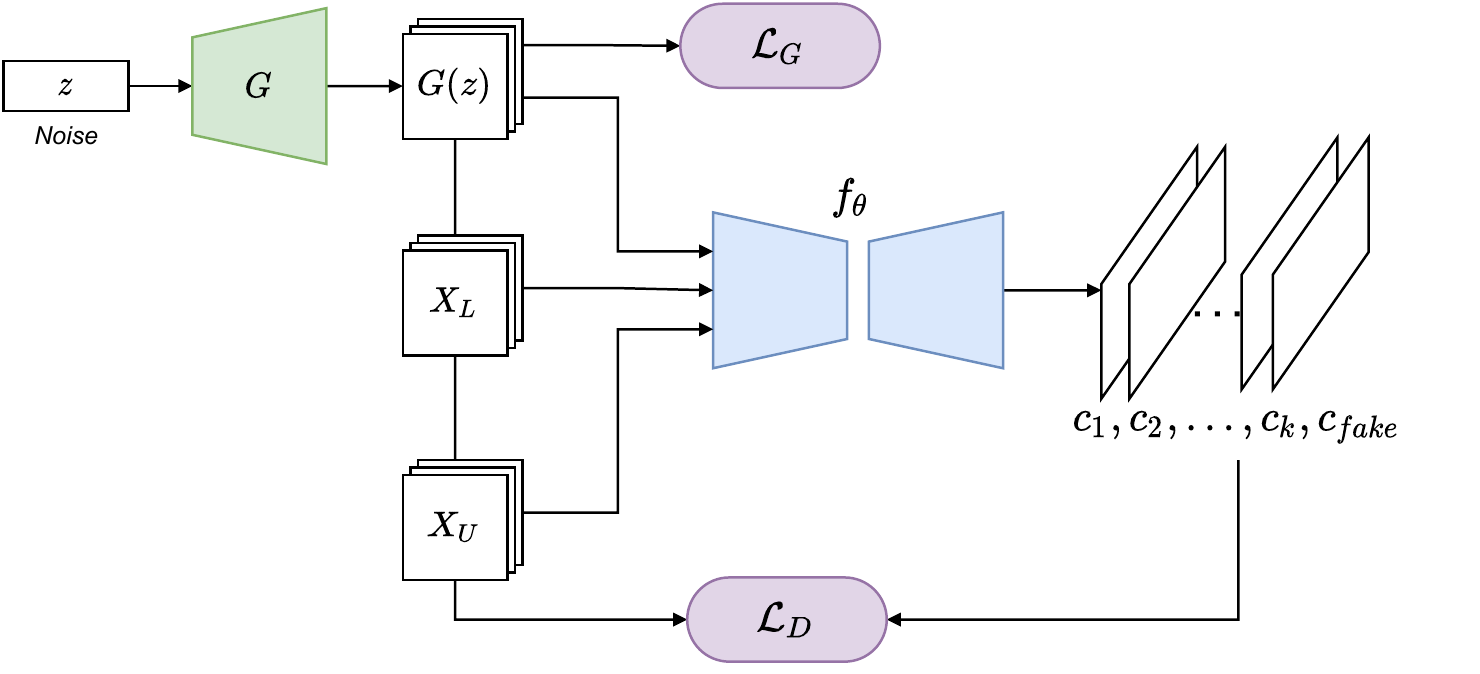}
  \caption{\small Generative adversarial method structure for semi-supervised segmentation. The generator $G$ receives random noise as input and generates new synthetic images. Then, the segmentation network $f_{\theta}$ receives both synthetic ($G(z)$) and real ($X_L, X_U$) images as input and classifies each pixel into its corresponding class $c_1,c_2,...,c_k$ or into an additional fake class $c_{fake}$ which indicates that it is a synthetic pixel. $\mathcal{L}_D$ and $\mathcal{L}_G$ are the discriminator and generator loss functions, respectively.}\label{fig: generative}
\end{figure}

\begin{figure}[h]
  \centering
  \includegraphics[width=0.47\textwidth]{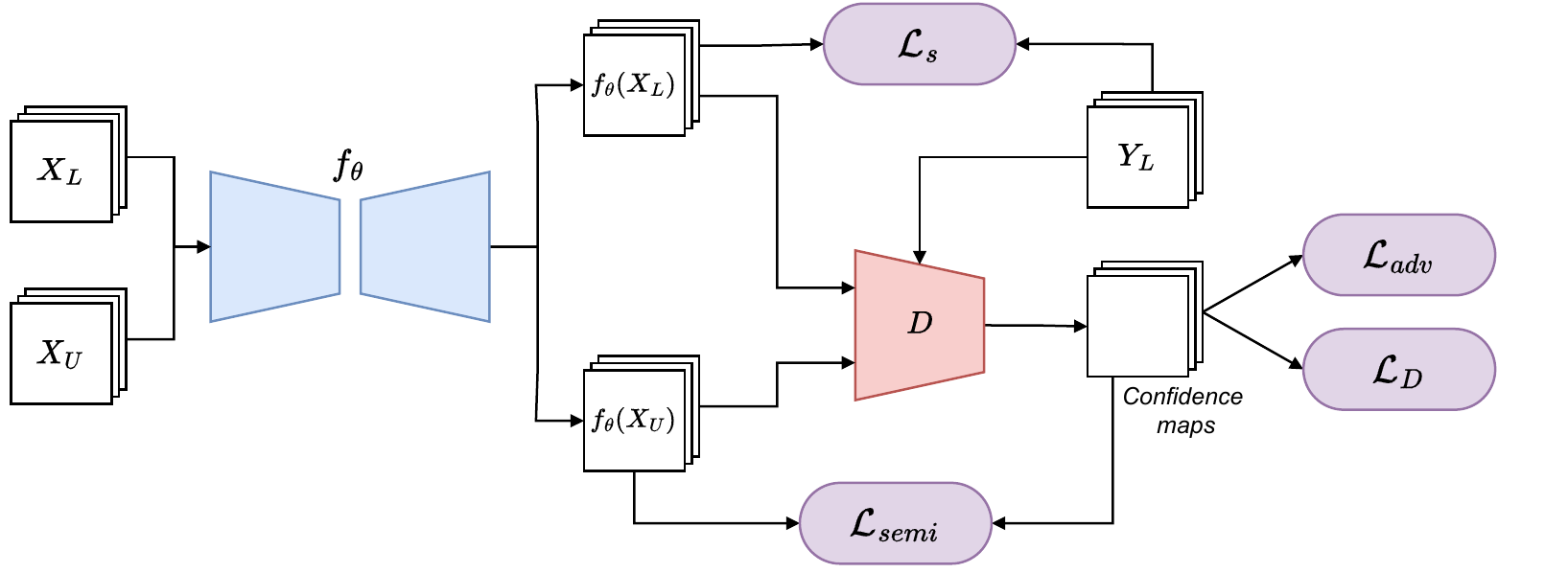}
  \caption{\small Non-generative adversarial method structure for semi-supervised segmentation. Segmentation network $f_{\theta}$ acts as generator. Supervised cross-entropy loss function ($\mathcal{L}_{sup}$) is used to train $f_{\theta}$ in a supervised way. Discriminator $D$ is trained to distinguish between real and predicted (by $f_{\theta}$) segmentation maps. The output of $D$ (\textit{confidence maps}) is used to perform the semi-supervised learning ($\mathcal{L}_{semi}$) with unlabeled data, and also is used for discriminator and adversarial loss functions ($\mathcal{L}_{D}$ and $\mathcal{L}_{adv}$). }\label{fig: non-generative}
\end{figure}

Methods based on adversarial training for semi-supervised SS are divided into two subcategories in the proposed taxonomy. The key aspect that differentiates these methods is the inclusion or non-inclusion of a generative model in the training process. In this sense, one of the subcategories groups those models that employ a generative model \cite{Souly2017SemiSS, Li2021SemanticSW}, thus generating new synthetic images that can be used as additional input examples for the segmentation model (Figure \ref{fig: generative}). On the other hand, the other subcategory groups those methods that do not include a generative model in their GAN-like structure \cite{Hung2018AdversarialLF,Mittal2021SemiSupervisedSS,di2021semi,liu2019semi,xu2021semi,Mendel2020SemisupervisedSB,Ke2020GuidedCT,Zhang2021StableSA,Jin2021AdversarialNI}. In these cases, a segmentation network assumes the role of generator, and the objective of the discriminator is differentiating those segmentation maps generated from the segmentation network from the real segmentation maps (i.e. ground truth) (Figure \ref{fig: non-generative}). Below we present and explain the different methods proposed in each of the two subcategories.

\subsubsection{Generative methods}

The methods based on GANs, in general, were the first approaches of DL techniques to the problem of semi-supervised SS. Previously, only weakly supervised approaches had been proposed, which do not take advantage of unlabeled data.

In particular, the first method \cite{Souly2017SemiSS} proposed to address the segmentation problem in a semi-supervised way, without requiring weak labels, consists of a GAN framework adapted for the segmentation problem. This framework aims, on the one hand, to handle and extract knowledge from a large amount of unlabeled data, and on the other hand, to increase the number of training examples available through the synthetic generation of images. Specifically, this method includes a generative network that approximates the distribution of the target images, thus achieving the ability to generate new training examples. A segmentation network assumes the role of discriminator and segments the images received as input, both real and synthetic. This network classifies each pixel with its corresponding class, or with an extra fake class, which indicates that this pixel or region of the image has been generated by the generator. This type of architecture can be seen represented in Figure \ref{fig: generative}. This approach adapts the loss function proposed for the original GAN to the SS problem. Both the loss function used to optimize the generator ($\mathcal{L}_G$) and the segmentation model that acts as a discriminator ($\mathcal{L}_D$) are shown below:

\begin{equation}\label{eq:ld_generative}
\small
\begin{split}
\mathcal{L}_D = - &\mathbb{E}_{x  \sim X}[log (f_{\theta}(x))] - \mathbb{E}_{z  \sim p_z(z)} [log(1-f_{\theta}(G(z)))] + \\ &\gamma \mathbb{E}_{x,y  \sim X_L}[CE(y,f_{\theta}(x))] 
\end{split}
\end{equation}
\begin{equation}\label{eq:lg_generative}
\small
\mathcal{L}_G = \mathbb{E}_{z  \sim p_z(z)} [log(1-f_{\theta}(G(z)))]
\end{equation}

The discriminator loss function $\mathcal{L}_D$ (Equation \ref{eq:ld_generative}) is composed of three terms. The first term penalizes the model when it labels a real sample as fake. The second term penalizes the model when it labels a fake sample as real. The last term is the supervised component, which tries to force the correct classification of each pixel of the labeled set in its corresponding class. $\gamma$ is the weight of the supervised component in the training process. The generator loss function $\mathcal{L}_G$ (Equation \ref{eq:lg_generative}), on the other hand, seeks to increase the quality of the generated images by penalizing $G$ when $f_{\theta}$ detects synthetic images.

Another generative method \cite{Li2021SemanticSW} has been proposed recently for semi-supervised SS, due to the recent success of StyleGAN \cite{Karras2020AnalyzingAI}. Specifically, the proposed model extends the StyleGAN model, adding a label synthesis branch, and attempts to capture the joint distribution of images and labels, gaining the ability to generate new image-label pairs. However, due to the high complexity of this generative problem, the authors themselves state that this approach is still far from being able to deal with the segmentation of natural and generic images, and limit its success cases to very specific domains such as skin lesions and facial parts segmentation.

\subsubsection{Non-generative methods}

On the other hand, we grouped those methods that use adversarial training and have a similar structure to GAN, but do not include a generative model. All the methods that we group under this subcategory share the characteristic of replacing the typical generative network of the classical GAN by a segmentation network. Its output is directed towards a discriminator that distinguishes between the real segmentation maps, and those generated by the segmentation network.

This GAN-like architecture for SS was originally proposed in \cite{Luc2016SemanticSU}, and adapted for a semi-supervised scenario in \cite{Hung2018AdversarialLF}. The authors present a fully convolutional discriminator that receives both segmentation maps (the one coming from the ground truth and the one predicted by the segmentation model, in this case DeepLabV2 \cite{Chen2018DeepLabSI}). The discriminative network is adversarially trained, together with the segmentation model, to distinguish real label maps from predicted ones. In this sense, it produces a probability map as output, of the same dimension as the input image, where it represents, for each pixel, the confidence of being a real example or a prediction made by the segmentation network. In this way, this confidence map indicates the quality of the segmentation in a certain area, so that the confidence map of the unlabeled images can be used to detect those areas where the predicted labels have enough quality to be used in the training process of the segmentation model. This structure is represented in Figure \ref{fig: non-generative}. The formulation of the loss functions involved in these methods is presented below:
\begin{equation}\label{eq:ld_non_generative}
\small
\mathcal{L}_D = - \mathbb{E}_{y  \sim X_L}[log (D(y))] - \mathbb{E}_{x  \sim X} [log(1-D(f_{\theta}(x)))] 
\end{equation}
\begin{equation}\label{eq:lseg_non_generative}
\small
\mathcal{L}_{seg} = \mathcal{L}_{sup} + \lambda_{adv} \mathcal{L}_{adv} + \lambda_{semi} \mathcal{L}_{semi}
\end{equation}
\begin{equation}\label{eq:lsup_non_generative}
\small
\mathcal{L}_{sup} = \mathbb{E}_{x,y  \sim X_L}[CE(y,f_{\theta}(x))]
\end{equation}
\begin{equation}\label{eq:ladv_non_generative}
\small
\mathcal{L}_{adv} = - \mathbb{E}_{x  \sim X}[log (D(f_{\theta}(x)))]
\end{equation}
\begin{equation}\label{eq:lsemi_non_generative}
\small
\mathcal{L}_{semi} = - \mathbb{E}_{x  \sim X_U} [I(D(f_{\theta}(x)) > \mathcal{T})  \cdot \hat{y} \cdot log(f_{\theta}(x))]
\end{equation}

The discriminator loss function $\mathcal{L}_D$ (Equation \ref{eq:ld_non_generative}) is composed of two terms, each of which forces the discriminator $D$ to detect the segmentation maps coming from the ground truth and those generated by the segmentation network $f_{\theta}$. The segmentation network loss function $\mathcal{L}_{seg}$ (Equation \ref{eq:lseg_non_generative}) is composed of three terms. The first is the supervised component $\mathcal{L}_{sup}$ (Equation \ref{eq:lsup_non_generative}), formed by the cross-entropy loss function. The second is the adversarial component $\mathcal{L}_{adv}$ (Equation \ref{eq:ladv_non_generative}) that penalizes the cases in which $D$ detects segmentation maps generated by the segmentation network. The third term $\mathcal{L}_{semi}$ (Equation \ref{eq:lsemi_non_generative}) allows to take into account the unlabeled images whose segmentation exceeds a confidence threshold $\mathcal{T}$ by $D$. $\lambda_{adv}$ and $\lambda_{semi}$ are parameters that weight the use of their respective terms.

Based on the previous approach, other alternatives have been proposed to improve the structure of the original method in different ways. S4GAN \cite{Mittal2021SemiSupervisedSS} proposes the use of a simpler discriminator that generates an output for the entire segmentation map rather than for each pixel. It also includes an additional processing branch where a classifier is trained. It is used to filter the segmentation maps obtained, removing those labels that are false positives in view of the classifier. Confrontation Network \cite{di2021semi} method also incorporates image-level discriminator and improves the generator loss function by adding a variance regularization term. Other approaches \cite{liu2019semi, xu2021semi} propose the use of two discriminators, one at the image level and the other at the pixel level. Both are used together in order to increase the accuracy in the definition of confidence areas in the images.

Error-Correcting Supervision (ECS) \cite{Mendel2020SemisupervisedSB} and Guided Collaborative Training (GCT) \cite{Ke2020GuidedCT} are based on a collaborative strategy, very close to the original adversary strategy. These approaches introduce a new network which assumes the role of discriminator, called correction network in the case of ECS and flaw detector in GCT. These approaches provide, in addition to a confidence map at the pixel level, a correction for those areas where confidence is low.

Other adversarial approaches incorporate attention modules with the objective of modeling long-range semantic dependencies. This is the case in \cite{Zhang2021StableSA} which also incorporates spectral normalization to reduce the instability in the training process. Another approach \cite{Jin2021AdversarialNI} proposes the use of attention modules in combination with  sparse representation module that helps the segmentation model to emphasize the edges and locations of objects.

\subsection{Consistency regularization}\label{subsec: consistency}

SSL makes some assumptions without which successful knowledge extraction from unlabeled data would not be possible. Specifically, consistency regularization methods are based on the assumption of smoothness \cite{Chapelle2006SemiSupervisedL}. This assumption says that, for two nearby points in the input space, their labels must be the same. In other words, a robust model should obtain similar predictions for both a point and a locally modified version of it. 
In this sense, SSL methods based on consistency regularization take advantage of unlabeled data by applying perturbations on them, and training a model that is not affected by these perturbations. This is achieved by adding a regularization term to the loss function that measures the distance between the original and perturbed predictions. The following is the formal definition of the described loss function:





\begin{equation}
\small
\mathcal{L} = \mathcal{L}_{sup} + \lambda \mathcal{L}_{cons}
\end{equation}
\begin{equation}
\small
\mathcal{L}_{sup} = \mathbb{E}_{x,y  \sim X_L}[CE(y,S(x))]
\end{equation}
\begin{equation}
\small
\mathcal{L}_{cons} = \mathbb{E}_{x  \sim X_U}[R(f_{\theta}(x), f_{\theta'}(x))]
\end{equation}

where $\mathcal{L}_{sup}$ is the supervised cross-entropy ($CE$) loss function and $\mathcal{L}_{cons}$ is the unsupervised regularization term. $R$ is a function that measure the distance between two predictions obtained from the student network $f_{\theta}$ and teacher network $f_{\theta'}$. $\lambda$ is used to weight the relevance of $\mathcal{L}_{cons}$.
\begin{figure}[h]
  \centering
  \includegraphics[width=0.47\textwidth]{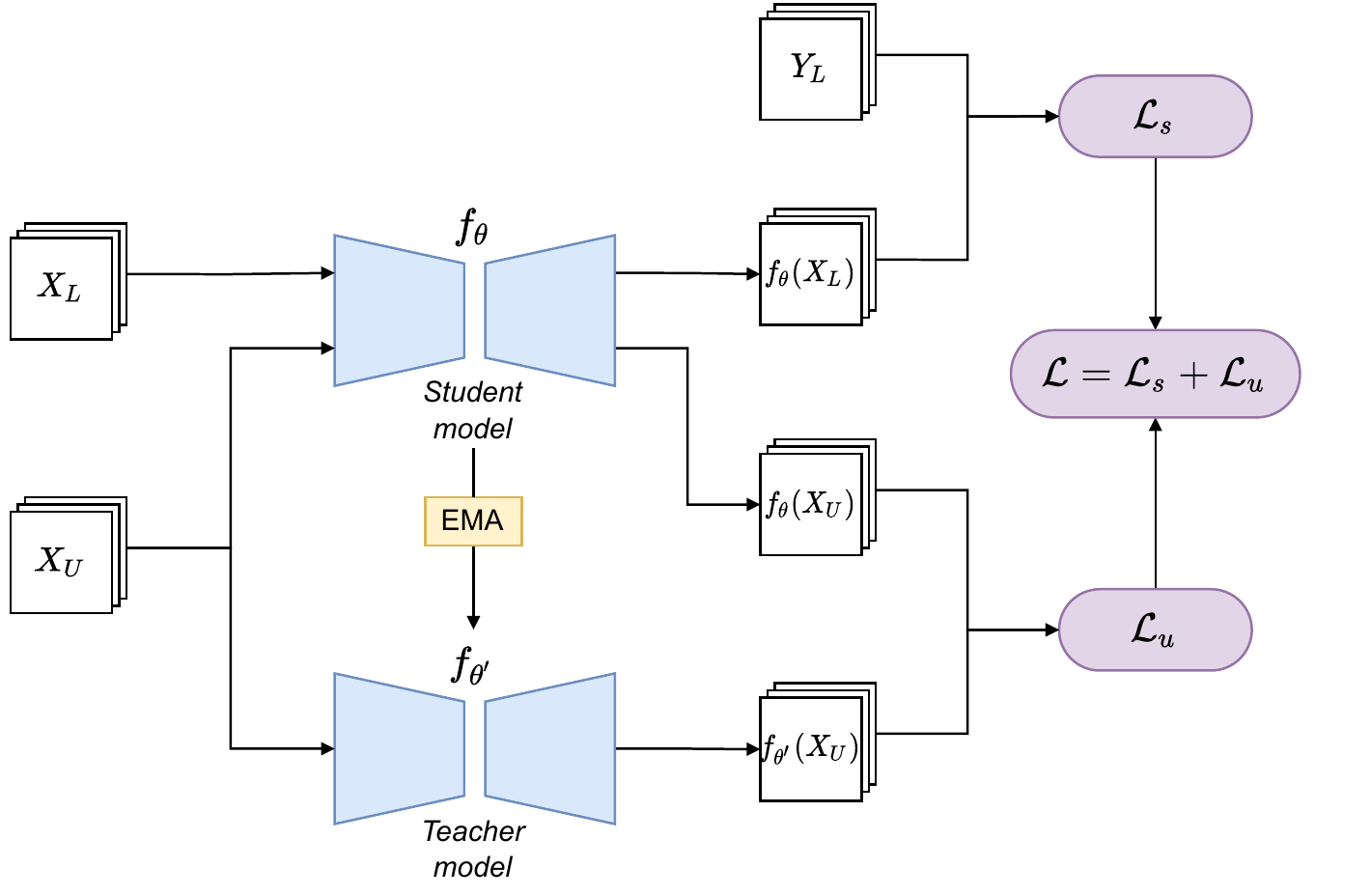}
  \caption{\small Mean Teacher \cite{Tarvainen2017MeanTA} method structure. $\mathcal{L}_{s}$ is used to train the student model $f_{\theta}$ in a supervised way. $\mathcal{L}_{u}$ is a regularization term that forces consistency between $f_{\theta}$ and teacher model $f_{\theta'}$ predictions.} \label{fig: mean}
\end{figure}

The basic method on which all other approaches are based is Mean Teacher \cite{Tarvainen2017MeanTA}. It forces consistency between the predictions of a student network and a teacher network. The weights of the teacher network are calculated by an exponential moving average (EMA) of the weights of the student network. Figure \ref{fig: mean} shows a graphical representation of the structure of this method.

\begin{figure}[h]
  \centering
  \includegraphics[width=0.47\textwidth]{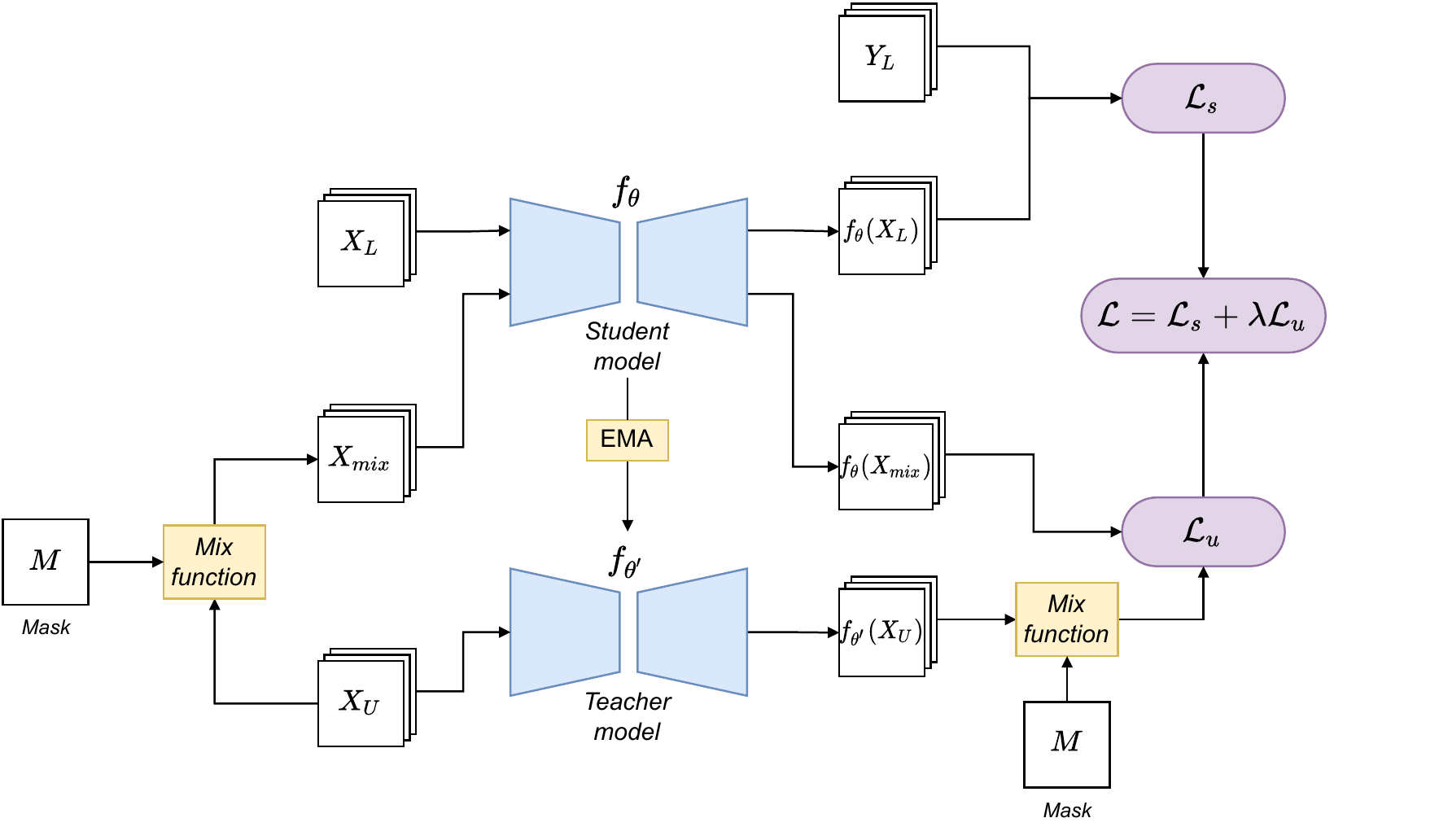}
  \caption{\small Input perturbations based consistency regularization method structure for semi-supervised segmentation. It presents a Mean Teacher base structure (see Figure \ref{fig: mean}), and incorporates input perturbations in the unlabeled data by means of the \textit{mix function} and the $M$ mask.}\label{fig: input}
\end{figure}

\begin{figure}[h]
  \centering
  \includegraphics[width=0.47\textwidth]{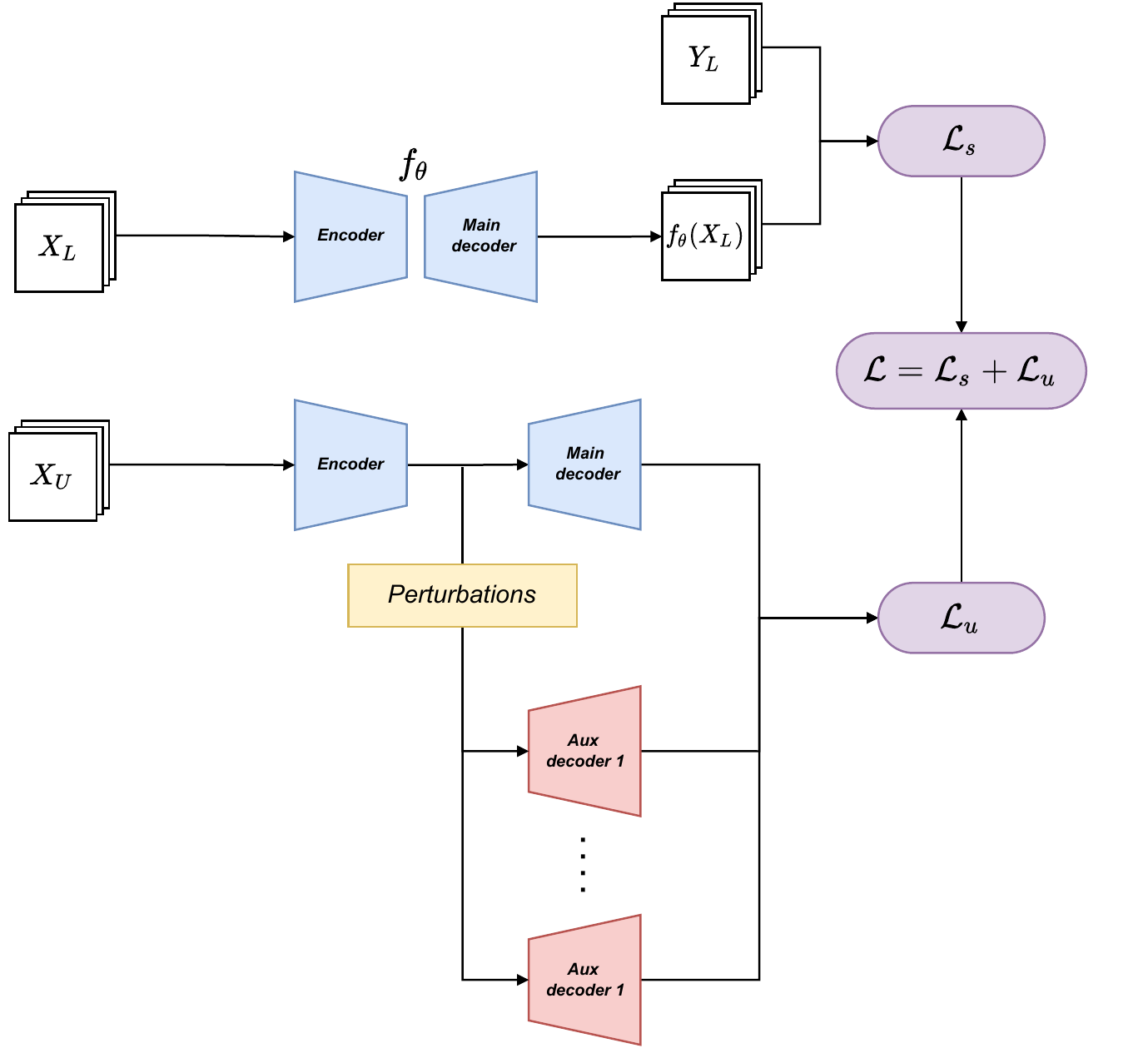}
  \caption{\small Feature perturbations based consistency regularization method structure for semi-supervised segmentation. It presents a Mean Teacher base structure (see Figure \ref{fig: mean}), and incorporates perturbations in an internal representation of the segmentation network, thus obtaining different outputs from auxiliary decoders. These outputs are forced to be consistent through the regularization term $\mathcal{L}_{u}$.}\label{fig: feature}
\end{figure}

\begin{figure}[h]
  \centering
  \includegraphics[width=0.37\textwidth]{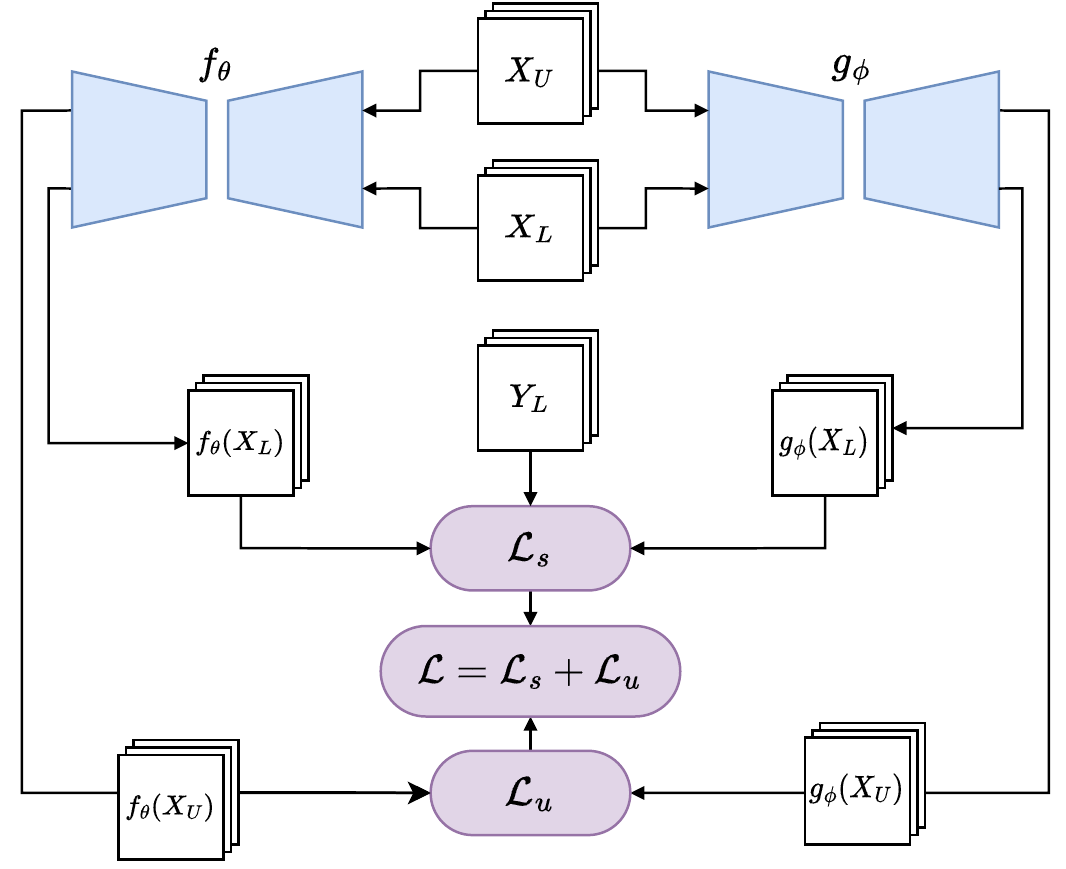}
  \caption{\small Network perturbations based consistency regularization method structure for semi-supervised segmentation. It presents a Mean Teacher base structure (see Figure \ref{fig: mean}) and changes the teacher model to a second segmentation network $g_{\phi}$ that is trained independently. The outputs of both networks are forced to be consistent by the regularization term $\mathcal{L}_{u}$.}\label{fig: network}
\end{figure}

The main difference between methods based on consistency regularization for semi-supervised SS lies in the way they incorporate perturbations to the data. Based on this, we can group these methods into four subcategories. On the one hand, the methods based on input perturbations \cite{French2020SemisupervisedSS, Olsson2021ClassMixSD, Kim2020StructuredCL, chen2021complexmix,li2020semi, grubivsic2021baseline}. These methods apply perturbations directly to the input images using data augmentation techniques. They force the model to predict the same label for both the original image and the augmented image (Figure \ref{fig: input}). Second, the methods based on feature perturbations, which incorporate perturbations internally in the segmentation network, thus obtaining modified features \cite{Ouali2020SemiSupervisedSS} (Figure \ref{fig: feature}). In third place, the methods based on network perturbations, which obtain perturbed predictions by using different networks, for instance, networks with different starting weights \cite{Chen2021SemiSupervisedSS,an2022deep,peng2020deep} (Figure \ref{fig: network}). Finally, we can identify a last subcategory that combines some of the three previous types of perturbations \cite{Liu2021PerturbedAS, wu2022perturbation}.

\subsubsection{Input perturbations}
In a first subcategory we group those consistency regularization methods that apply perturbations directly to the unlabeled input images using data augmentation techniques. Then, these methods train a segmentation model that is not sensitive to these input perturbations, and predicts segmentation maps that are as similar as possible for both the original images and their augmented versions. The key aspect that differentiates these methods is the way they perform modifications to the data. We can find in the literature different proposals for data augmentation techniques that have been applied to the semi-supervised SS problem. The consistency term incorporated in these data augmentation-based methods is defined as follows:
\begin{equation}
\small
\begin{split}
\mathcal{L}_{cons} = \mathbb{E}_{x_a,x_b  \sim X_U}[R(&mix(f_{\theta'}(x_a), f_{\theta'}(x_b),M)  ,\\ &f_{\theta}(mix(x_a,x_b,M)))]
\end{split}
\end{equation}

where $mix$ is a mixing function that receives as input two images $x_a,x_b$ (or segmentation maps $f_{\theta'}(x_a), f_{\theta'}(x_b)$) and returns a combination of them. This combination is done by means of a predefined mask $M$.
Below we detail the different data augmentation techniques for semi-supervised SS proposed in the literature. 

    
    CutOut and CutMix techniques are applied to SS in \cite{French2020SemisupervisedSS}. Previously, these techniques have been applied in image classification \cite{Devries2017ImprovedRO, Yun2019CutMixRS}. These techniques use a rectangular mask over the images. CutOut discards the rectangular section marked by the mask in the training process. Then, the consistency between the predictions of the original image and the modified image is forced by the regularization term. On the other hand, CutMix combines two images using a rectangular mask, obtaining a new image where the sections marked by the mask belong to one of the original images, and the rest of the sections belong to the other image (the inverse image is also obtained). Another approach \cite{Kim2020StructuredCL} extends the previous method by adding a new term to the loss function called consistency structured loss that incorporates the concept of pair-wise knowledge distillation \cite{Liu2019StructuredKD}.
    
    
    ClassMix \cite{Olsson2021ClassMixSD} is proposed and designed specifically for the SS problem. This technique differs from the previous CutMix technique in the form of the mask that is applied to mix images. In this case, the sections marked by the mask coincide with areas belonging to the same class in the image, so that sections completely belonging to one class are copied into another image, thus generating the new augmented images. The difference between original and augmented predictions is calculated in the same way as the previous technique using the regularization term. ComplexMix \cite{chen2021complexmix} proposes the combined use of the previous data augmentation techniques, CutMix and ClassMix.
    
    Besides these types of methods that propose a specific data augmentation technique for segmentation, other approaches \cite{li2020semi} use classical data augmentation techniques (e. g. cropping, color jittering or flipping) to obtain the perturbed versions of the original images. Focused on efficiency, a method \cite{grubivsic2021baseline} is proposed that performs photometric and geometric perturbations only in the teacher model.


\subsubsection{Feature perturbations}

The second way to introduce perturbations in the training process consists in perturbing the internal features of the segmentation network. Cross-Consistency Training (CCT) \cite{Ouali2020SemiSupervisedSS} is proposed to address the semi-supervised SS problem following this idea. The architecture presented extends a supervised segmentation model with an encoder-decoder structure (e.g. DeepLabV3+ \cite{Chen2018EncoderDecoderWA}) with some auxiliary decoders. First, a supervised training is carried out with the available labeled data, using the main decoder. Next, to take advantage of the unlabeled data, the encoder output is perturbed in different ways, resulting in different versions of the same features, which are directed to different auxiliary decoders. Finally, consistency between the outputs of the auxiliary decoders is enforced, favoring similar predictions for different perturbed versions of the encoder output features. The consistency term incorporated in these feature perturbation-based methods is defined as follows:
\begin{equation}
\small
\mathcal{L}_{cons} = \mathbb{E}_{x  \sim X_U}[ \frac{1}{k} \sum_{k=1}^K R(h(x), h^k(x))]
\end{equation}

where $h$ is the main decoder, $h^k$ is the k-th auxiliary decoder, and $K$ is the number of auxiliary decoders.

\subsubsection{Network perturbations}

Another way of introducing perturbations in the training process is to use different segmentation networks. The differences between the networks constitute the perturbations in the resulting predictions. This is the case of the Cross Pseudo Supervision (CPS) method \cite{Chen2021SemiSupervisedSS}, which follows a training process similar to Mean Teacher. In this case the training of the two networks involved is carried out in a parallel and independent way, instead of updating one according to the EMA of the other. In addition, although both networks share the same architecture, they are initialized with different random weights, thus increasing the difference between them. An extension of the above method by including three networks in the training process can be seen in \cite{an2022deep}. Another approach \cite{peng2020deep} emphasizes the importance of enforcing diversity across networks and proposes the use of adversarial samples and re-sampling strategy to train the models on different sets. 

As in the other consistency regularization methods, the consistency between the predictions of the networks involved for unlabeled images is enforced by a regularization term included in the loss function. This regularization term is defined as follows (for the case where two networks are used):

\begin{equation}
\small
\mathcal{L}_{cons} = \mathbb{E}_{x  \sim X_U}[R(f_{\theta}(x), g_{\phi}(x))]
\end{equation}

where $f_{\theta}$ and $g_{\phi}$ are different networks trained independently.

\subsubsection{Combined perturbations}

Finally, a last subcategory includes those methods that jointly apply several of the different types of perturbations described above. 

A method that proposes the combination of input, feature, and network perturbations is presented in \cite{Liu2021PerturbedAS}. This method emphasizes the fact that a greater variety and strength of perturbations may cause more problems if the predictions are not sufficiently accurate. In this sense, to ensure accurate predictions for unlabeled images, this method extends the Mean Teacher method by adding a confidence-weighted cross-entropy loss function, instead of the mean square error (MSE) used by the classic Mean Teacher method. In addition, it also proposes a new way of performing feature perturbations by means of virtual adversarial training \cite{Miyato2019VirtualAT}.

The combination of input perturbations, specifically the CutMix technique, and feature perturbations is proposed in \cite{wu2022perturbation}. Instead of adding different auxiliary decoders, as in CCT \cite{Ouali2020SemiSupervisedSS} , this method proposes the application of perturbations directly on the features, while the decoders share the weights.


\subsection{Pseudo-labeling methods} \label{subsec: pseudo-labeling}

Pseudo-labeling methods, also known as bootstrapping \cite{Ouali2020AnOO}, wrapper \cite{vanEngelen2019ASO} or self-labeled \cite{Triguero2013SelflabeledTF} methods, are among the most widely known and the first semi-supervised methods to appear \cite{xiaojin2008semi}. This type of method consists of an intuitive approach to extend existing supervised models to a semi-supervised scenario, allowing them to handle unlabeled data. The idea behind pseudo-labeling methods is simple: generate pseudo-labels of the unlabeled images from the predictions made by a model previously trained on the labeled data. Then, extend the labeled dataset with these new pairs of images and pseudo-labels, and train a new model on this new dataset. This idea is formalized with the following definition of loss function:
\begin{equation}
\small
\mathcal{L} = \mathbb{E}_{x,y  \sim X_L}[CE(y,f_{\theta}(x))] + \lambda \mathbb{E}_{x  \sim X_U}[CE(\hat{y},f_{\theta}(x))]
\end{equation}

where $\hat{y}$ is the pseudo-label for image $x$, generated from the predicted probabilities with the segmentation model $f_{\theta}$, in many cases by one-hot encoding, and $\lambda$ is a parameter that weights the unsupervised part of the loss function.

\begin{figure}[h]
  \centering
  \includegraphics[width=0.47\textwidth]{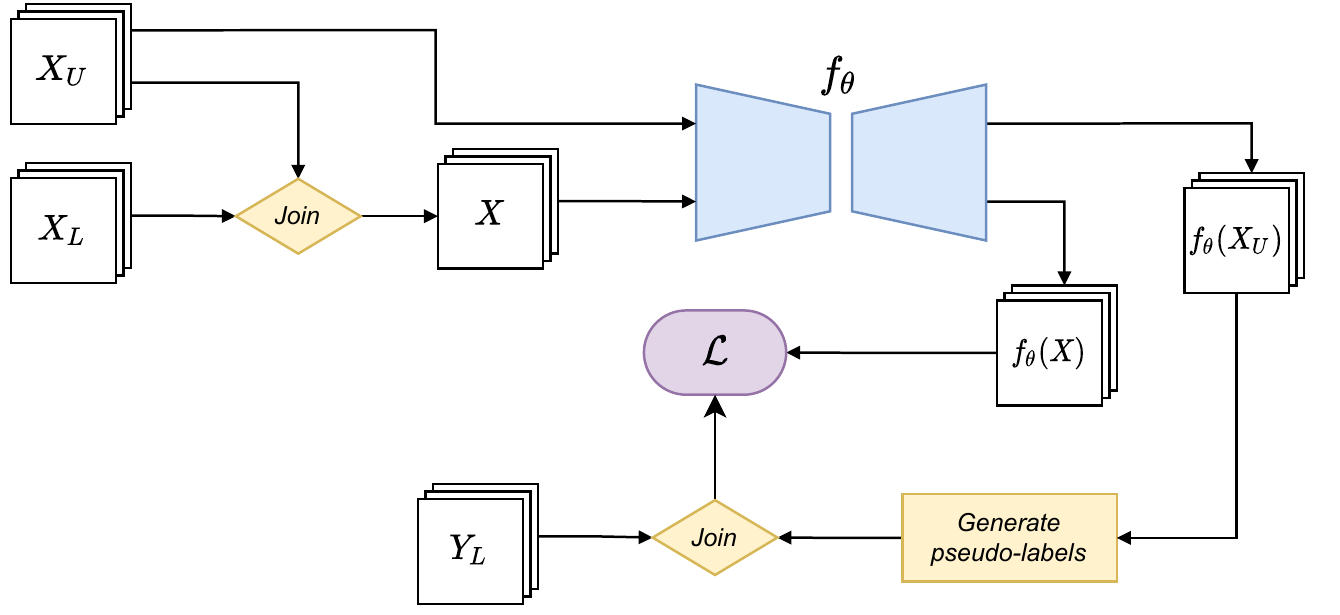}
  \caption{\small Self-training method structure for semi-supervised segmentation. Firstly, pseudo-labels are generated for the unlabeled images using the segmentation network $f_{\theta}$ (usually pre-trained with labeled images). Then, pseudo-labels are joined to the ground truth and the loss function $\mathcal{L}$ is computed in a supervised way for all images.} \label{fig: self-training}
\end{figure}
\begin{figure}[h]
  \centering
  \includegraphics[width=0.47\textwidth]{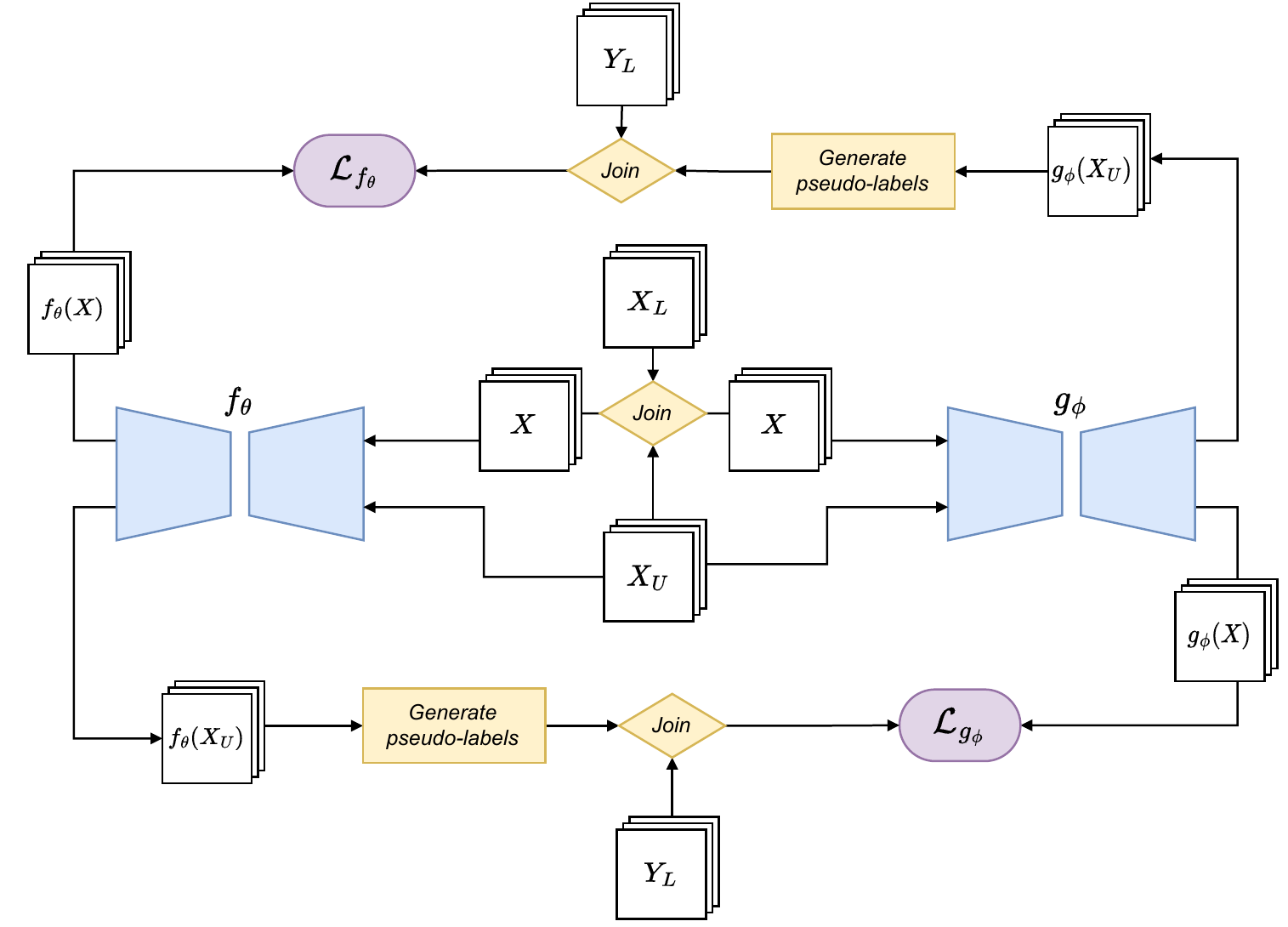}
  \caption{Mutual-training method structure for semi-supervised segmentation. This approach extends the classical self-training (see Figure \ref{fig: self-training}) with an additional segmentation network $g_{\phi}$. The pseudo-labels used to retrain each of the networks are computed with the other network.}\label{fig: mutual-training}
\end{figure}

Based on the differences between models involved in the training process and the way pseudo-labels are generated, in our taxonomy we differentiate between two types of pseudo-labeling methods. The first are self-training methods \cite{Yang2021STMS, Teh2021TheGA, Zhu2020ImprovingSS, li2021residual, chen2020digging, Yuan2021ASB, he2021re}, based only on one supervised base model and representing the simplest form of pseudo-labeling, where pseudo-labels are generated from their own high-confidence predictions (see Figure \ref{fig: self-training}). Secondly, mutual-training methods \cite{feng2020dmt, zhou2022catastrophic}, which involve multiple models with explicit differences such as different initialization weights or training on different views of the dataset. Each of the models are retrained with the unlabeled images and the corresponding pseudo-labels generated by other models involved in the process (see Figure \ref{fig: mutual-training}).

\subsubsection{Self-training}

Self-training methods are the simplest pseudo-labeling and semi-supervised methods, first proposed in \cite{Yarowsky1995UnsupervisedWS}, thoroughly reviewed in \cite{Triguero2013SelflabeledTF} and applied for the first time with deep neural networks in \cite{Lee2013PseudoLabelT}. These methods consist in retraining a base supervised model by feeding back the training set with its own predictions. The typical self-training process consists of the following steps:
\begin{enumerate}
    \item The supervised model is trained on the available labeled data.
    \item Predictions are obtained from the unlabeled data using the previously trained model. Those predictions with a confidence level higher than a predefined threshold become pseudo-labels for unlabeled data and are included in the labeled data set.
    \item The supervised model is retrained with this new data set composed of the labeled and the pseudo-labeled data.
\end{enumerate}

This process can be repeated in an iterative way, obtaining new pseudo-labels with the model resulting from step 3, refining the quality of the pseudo-labels at each iteration, until no prediction exceeds the confidence threshold necessary to be treated as a pseudo-label. 

The methods grouped in this subsection are based on this training process applied to the SS problem, each of them contributing some variant to the original algorithm that improves the learning capacity. For instance, the method proposed in \cite{Zhu2020ImprovingSS} extends the original self-training process with a centroid sampling technique. The purpose is to solve the problem of class imbalancement in the pseudo-labels.


Other proposals consist of adding some auxiliary network to the self-training process. For example, in \cite{li2021residual} the authors extend the self-training process by adding a residual network. This network is trained with the labeled images, and is subsequently used to refine the pseudo-labels obtained by the segmentation model. The pseudo-labels predicted by a model may have a substantially different label space than the ground truth. This can be a problem when training a model with both label inputs, since it can lead to different gradient directions, resulting in a chaotic back-propagation process. A possible solution proposed in \cite{chen2020digging} consists in the use of a segmentation model that shares the encoder (i.e. ResNet101) and incorporates two different decoders, one for each label space.

The integration of data augmentation techniques within the self-training process has also been proposed in different approaches. The ST++ \cite{Yang2021STMS} method applies data augmentation techniques on the unlabeled images during the self-training process. This is combined with a selective stage in which, on each iteration of the self-training process, those images with reliable pseudo-labels are prioritized, and those images that present a higher probability of suffering from errors in the pseudo-labels are discarded.


Nevertheless, the application of data augmentation may alter the distribution of the mean and variance in the batch normalization. To solve this problem, the use of distribution-specific batch normalization is proposed in \cite{Yuan2021ASB}. Additionally, this method also integrates a self-correction loss function which performs a dynamic re-weighting based on confidence, in order to avoid over-fitting noisy labels and under-learning of the most difficult classes.

A common issue faced by this type of methods is the distribution mismatch between ground truth and pseudo-labels, where the latter are often biased towards the majority classes. In order to obtain unbiased pseudo-labels, a strategy of distribution alignment and random sampling with class-wise thresholding is proposed in \cite{he2021re}, also in combination with data augmentation techniques. 

Another proposal focuses on the difficulty of defining an optimal ratio between the actual labeled data and the pseudo-labeled data to be used in the self-training process. In this sense, two strategies are proposed to approach this optimal value during the iterative retraining process, one of them is based on a randomized search (RIST) and the other one employs a greedy algorithm (GIST) \cite{Teh2021TheGA}. 

\subsubsection{Mutual-training}

One of the main disadvantages of previously described self-training methods is the absence of a mechanism for detecting their own errors. Instead of learning from their own predictions, mutual learning \cite{Zhang2018DeepML} methods extend self-training methods and involve multiple learning models, each of which train with the pseudo-labels generated by other models. The diversity present among the participating models is one of the key aspects for the proper performance of this type of methods \cite{Wang2010ANA}. That is why the different existing proposals try to explicitly induce differences between the base supervised models that compose the co-training method, for instance, by initializing such models with different pre-trained weights or by training each of the models with different views or subsets of the training set. In other studies, similar methods have been categorized as disagreement-based \cite{Zhou2008SemisupervisedLB, Yang2021ASO}, since they rely primarily on exploiting the predictive differences between the models involved, multi-view training \cite{Ouali2020AnOO} or co-training \cite{vanEngelen2019ASO}.

Dynamic Mutual Training (DMT) is a mutual learning approach adapted to semi-supervised scenario and SS problem proposed to take advantage of the disagreement between models as a way to detect errors in the generated pseudo-labels. This method takes these differences into account by means of a loss function that is dynamically re-weighted during training based on the discrepancies between two different models, which are trained independently, using the pseudo-labels generated by the other model. In this sense, a greater disagreement in a specific pixel indicates a greater probability of error, so it is weighted with a low value in the loss function, and has less influence on the training than other pixels or areas of the image where the discrepancy between models is smaller \cite{feng2020dmt}.

Another approach consists in extending the previous method (DMT) with a pseudo-label enhancement strategy \cite{zhou2022catastrophic}. This publication focuses on the problem of catastrophic forgetting. This problem points out the difficulty that models have to maintain the acquired knowledge when they receive inputs with some variants. This could be the case of pseudo-labels. In order to maintain the acquired knowledge during the whole training process, and to avoid that the models suffer a bias towards the last classes learned, the authors propose a strategy that takes into account the pseudo-labels generated in previous stages to refine the current ones.

\subsection{Contrastive learning} \label{subsec: contrastive}

Contrastive learning focuses on high-level features to differentiate between classes in the absence of ground truth. In other words, these types of methods group similar samples and move them away from different samples in feature space. In many contrastive learning methods, the target sample to be compared is called the \textit{query}, while the similar and dissimilar samples are called the \textit{positive}  and \textit{negative} keys, respectively. Due to the lack of annotations in the data, samples considered similar in the training process are augmented versions of the same sample, while the rest of the data are considered different samples. Specifically, in the most relevant contrastive methods, pairs of augmented images are commonly obtained in different ways. Some of them apply data augmentation techniques (e. g. cropping, color jittering or flipping) as in the SimCLR method \cite{Chen2020ASF, Chen2020BigSM}. Other methods divide the image into different overlaying sub-patches and considering these patches independent images as in the CPC method \cite{Oord2018RepresentationLW}.

Due to the success of this type of methods, even outperforming its supervised counterpart in some specific problems, such as Pascal VOC object detection \cite{Falcon2020AFF}, in recent years a series of contrastive learning methods specifically designed for SS have been proposed. The ReCo method \cite{Liu2021BootstrappingSS} is one of the first contrastive learning proposals for SS. This method consists in chaining on top of the segmentation model encoder an auxiliary decoder that maps the input feature to a higher dimensional representation space, in which the sampling of queries and keys is carried out. By means of the proposed contrastive loss function the query is enforced to be close to the positive key in the representation space, and away from the negative key. Because using all pixels of a high-dimensional image to compute the contrastive loss function is impractical, ReCo method incorporates an active sampling strategy that samples less than 5\% of the total pixels in the image. On the one hand, this method gives a higher probability of being selected as key negative those pixels belonging to classes that are usually confused with the query class. On the other hand, it relies on prediction confidence to select those pixels that are more difficult to classify for the segmentation model as query pixels.

Another contrastive learning method proposed for semi-supervised SS is based on positive-only contrastive learning \cite{Chen2021ExploringSS}, in which only positive keys are sampled. The key element of this method is the creation and dynamic updating of a memory bank containing a subset of samples from the labeled set. The samples with a higher prediction confidence are selected to be stored. Subsequently, a contrastive loss function ensures that the features of a sample are close to the features of the samples of the same class stored in the memory bank \cite{Alonso2021SemiSupervisedSS}. 


\subsection{Hybrid methods} \label{subsec: hybrid}


The last category includes those methods that share characteristics of several of the previously introduced categories. Hybrid methods that attempt to take advantage of the benefits of pseudo-labeling and consistency regularization methods are some of the most common in this category. For instance, a three-stage self-training framework whit an intermediate stage of consistency regularization \cite{Ke2022ATS} is proposed. Specifically, a multi-task model is integrated in the self-training process. It is trained on the segmentation problem using consistency regularization (task 1), and statistical information is introduced into the optimization process from the pseudo-labels (task 2).


In the same way, Adaptive Equalization Learning (AEL) \cite{Hu2021SemiSupervisedSS} also incorporates characteristics of consistency regularization and pseudo-labeling methods. AEL method is based on FixMatch \cite{Sohn2020FixMatchSS}, a widely used hybrid method originally proposed for image classification. It is common in segmentation problems that models underperform in some classes, mainly due to their difficulty or negative imbalance with respect to the rest of the classes. AEL focuses on these challenging classes. This method proposes a confidence bank that dynamically stores the performance of each category during training. Data augmentation techniques and adaptive equalization sampling are used to favor the training towards those disadvantaged classes.

Pseudo-Seg \cite{Zou2021PseudoSegDP} also integrates characteristics of consistency regularization and pseudo-labeling methods. The authors emphasize the fact that the usual ways of obtaining pseudo-labels (from the outputs of a trained segmentation model and applying a confidence threshold) can fail and result in low-quality pseudo-labels. To address this problem, an approach focused on performing a structured and quality design of pseudo-labels is proposed. This method generates the pseudo-labels from two different sources: on the one hand, the output of the segmentation model and, on the other hand, the output of a class activation map algorithm \cite{Selvaraju2019GradCAMVE}. Unlike the segmentation task that seeks to obtain a dense and accurate prediction, the class activation algorithms perform a simpler task in which they only need to predict coarser-grained outputs.

A key bottleneck in semi-supervised segmentation methods can be to treat labeled and unlabeled data separately during training. This is the issue that the hybrid GuidedMix-Net method focuses on \cite{Tu2021GuidedMixNetLT}, allowing a transfer of knowledge from labeled to unlabeled images. This is achieved through an interpolation between pairs of labeled and unlabeled images, thus capturing interactions between them.



Interest in methods that combine consistency regularization with contrastive learning has also increased recently. In this line, methods such as directional context-aware (DCA) \cite{Lai2021SemisupervisedSS} have been proposed. The authors point out the difficulty of generalizing in a semi-supervised environment, where the contexts of a given object are limited in the reduced set of labeled images. This may cause a segmentation model to give too much importance to these specific contexts, not focusing on some important characteristics of the object to be segmented.  To address this issue, The DCA method incorporates a new data augmentation technique that makes two cuts of the same image with an overlapping region. In this way it simulates two different contexts for that region, and enforces consistency between the two slices by means of a contrastive loss function.

The approach proposes in \cite{Zhong2021PixelCS} tries to achieve the same two properties: consistency in the prediction space and contrastiveness in the feature space. On the one hand, they enforce consistency between the predictions of two augmented versions of an unlabeled image using the $l_2$ loss. On the other hand, they integrate contrastive learning by means of a contrastive loss function that brings positive (similar) pairs closer and negative (dissimilar) pairs away in the feature space.

Another method that combines consistency regularization and contrastive learning is C3-SemiSeg, presented in \cite{zhou2021c3}. In this method, consistency regularization is focused on exploiting feature alignment under perturbations, introducing a novel cross-set region-level data augmentation strategy. In addition, cross-set contrastive learning is integrated to improve the feature representation capability.

A method presented in \cite{xiao2022semi} combines a consistency regularization framework based on cross-teacher training (CCT) with two complementary contrastive learning modules. CCT framework reduces the error accumulation between teacher and student networks while the contrastive learning modules promote class separation in the feature space. 

Finally, a method combining consistency regularization and adversarial training has been recently proposed \cite{cao2022adversarial}. In this case, a data augmentation technique that tries to maintain the image context is proposed. Additionally, a new adversarial dual-student framework is proposed in order to improve the performance of the classical Mean Teacher.

\section{Experimental setup}
\label{sec:experiments}


The main obstacle to have a realistic perception of the performance of the different state-of-the-art methods is the non-homogeneity of the comparative experiments presented. As a consequence, a direct comparison of the results obtained by each method is impossible. Among these differences we can find the use of different datasets or partitions of labeled and unlabeled data, different base models on which semi-supervised methods are based or different preprocessing or data augmentation techniques.

That is why the main goal of this experimental section is to offer the reader a comparison with unified, fair and equal conditions for all methods, thus offering a quick and accessible way to know the actual state-of-the-art methods in the field and their quality in comparison with others. To this end, we have carried out a series of experiments taking into account some guidelines that try to eradicate the comparison problems described above, on a selection of methods that tries to be representative for all the categories introduced in our taxonomy.

Our experimentation is mainly conducted in two directions. On the one hand, we propose an experiment with exhaustive representation of all categories of methods, on a range of partitions with different ratios of labeled and unlabeled data, with the aim of having quantitative results that allow a direct and fast comparison between the performance of the different methods. On the other hand, we propose another experimentation with some of the most relevant methods in the literature to perform a qualitative and visual comparison of the results obtained.


\hphantom{~~~~~}\textbf{Datasets.} 
For each experiment described in the previous section we chose a dataset that we consider to have the necessary characteristics to carry out the desired comparison. A detailed description of the following datasets can be found in section \ref{subsec:datasets}. We employ the PASCAL VOC 2012 \cite{Everingham2009ThePV} dataset in the experiment related to the quantitative comparison of state-of-the-art methods. This dataset is the most commonly used in the semi-supervised SS literature. In addition, it has a high number of images which helps to have stability in the results obtained. Second, for qualitative and visual comparison of the results we considered the Cityscapes dataset \cite{Cordts2016TheCD}. This dataset has higher resolution images, which allows a better visualization. In addition, each of the images in this dataset has representation of many of the classes (unlike PASCAL VOC 2012, where each image focuses on one or a small number of classes). This allows us to see how the trained models perform in situations where there are adjacent areas of several similar classes or with semantic dependencies between them. These two features make Cityscapes an ideal dataset to perform visual and qualitative analysis.



\textbf{Partition protocol.} As discussed above, partitions of labeled and unlabeled data is a key aspect to take into account in semi-supervised experiments. In order to obtain comparable results with other experimental studies, it is important to use the same or similar data partitions. That is why in our experimentation we decided to use the partitions proposed in one of the most recent studies, which present a wide variety of scenarios in terms of labeling ratio. These partitions can be found at \footnote{\href{https://github.com/charlesCXK/TorchSemiSeg}{https://github.com/charlesCXK/TorchSemiSeg}}, and a detailed description of them is presented in Table \ref{tab: partitions details}.


\begin{table}[!htp]\centering
\caption{ \small Labeling ratio and number of labeled and unlabeled images in the proposed partitions. }\label{tab: partitions details}
\scriptsize
\resizebox{.47\textwidth}{!}{
\begin{tabular}{|c|c|c|c|c|}\hline
\textbf{Dataset} &\textbf{Labeling ratio} &\textbf{Labeled images} &\textbf{Unlabeled images} \\\hline \hline
\multirow{4}{*}{PASCAL VOC 2012 \cite{Everingham2009ThePV}} &1/100 &106 &10476 \\
 &1/50 &212 &10370 \\
  &1/20 &529 &10053 \\
   &1/8 &1323 &9259 \\\hline
Cityscapes \cite{Cordts2016TheCD} &1/8 &372 &2603 \\
\hline
\end{tabular}}
\end{table}

\textbf{Validation strategy.} The standard validation strategy in SS on the datasets used in our experiments consists in a simple hold-out, with a training set and a validation set. For each of the datasets, the composition of training and validation partitions is standard in the SS literature, so we use these same partitions for better generality of the analysis results. In the case of PASCAL VOC 2012, the training set is composed of 10582 images while the validation set is composed of 1449. For Cityscapes, the training set is composed of 2975 images while the validation set is composed of 500. 

\textbf{Performance metric.} The performance metric used in this experimentation, standard in the SS literature, is the mean intersection over union (mean IoU). Unlike accuracy, this metric tries to be robust to the presence of imbalanced classes, which is very common in problems where we have pixel-level labels as is the case in SS. Specifically, this metric computes the ratio between the number of true positives and the sum of true positives, false negatives and false positives, for each of the classes and averages these values.
\begin{equation}
\small
    mean IoU = \frac{1}{N} \sum_{i=1}^{N} \frac{N_{ii}}{ \sum_{j=1}^{N} N_{ij} + \sum_{j=1}^{N} N_{ji} - N_{ii}}
\end{equation}

where $N$ is the number of classes, $N_{ii}$ is the numbers of true positives for class $i$, $N_{ij}$ is the numbers of false positives for class $i$ and $j$ and $N_{ji}$ is the number of false negatives for class $j$ and $i$.

\textbf{Selection of state-of-the-art methods.} We include in the experimental study state-of-the-art methods such that all categories and subcategories defined in the taxonomy presented in section \ref{sec:taxonomy} are sufficiently covered. The main criteria taken into account when choosing a method from each category have been popularity of the method, in terms of number of citations, and availability of code. As baseline methods, we include the DeepLabV3 \cite{Chen2018EncoderDecoderWA} supervised model, trained only with the labeled partition, which is the base model for the rest of the semi-supervised methods, and the Mean Teacher \cite{Tarvainen2017MeanTA} method, which has a strong influence on most of the proposed methods. The s4GAN \cite{Mittal2021SemiSupervisedSS} method is included as an adversarial method. As methods based on consistency regularization we include the ClassMix \cite{Olsson2021ClassMixSD} method as an example of data augmentation perturbations, the CCT \cite{Ouali2020SemiSupervisedSS} method as a feature perturbation method, and the CPS \cite{Chen2021SemiSupervisedSS}  method for network perturbations. The ST \cite{Yang2021STMS} and DMT \cite{feng2020dmt} methods represent the pseudo-labeling-based methods, based on self-training and mutual-training, respectively. For contrastive learning we include the ReCO \cite{Liu2021BootstrappingSS} method, and finally as a hybrid method we include the CAC \cite{Lai2021SemisupervisedSS} method.

\textbf{Base model and backbone.} All semi-supervised methods for SS work by supporting a supervised segmentation model. The good performance of the semi-supervised method depends to a large extent on the base model. This is why the choice of this base model is critical. As well, segmentation models rely on a network (i.e., backbone), on which the final performance of the semi-supervised segmentation method also depends. The fact that the different proposals for semi-supervised methods rely on different base models and backbones makes it difficult to compare their performance, which is why in our experimentation we unified this critical aspect. We opt for DeepLabV3+ as the base model and ResNet101 as the backbone, this combination being one of the best performing in the literature.

\textbf{Hardware and software setup.} The entire experimental code has been developed using Python as programming language and PyTorch as Deep Learning framework. The different experiments have been run on a Tesla V100 GPU.

\section{Results and discussion}
\label{sec:results}

In this section we show and discuss the results obtained. First, in subsection \ref{subsec: quantitative} we present and discuss the quantitative results obtained on the PASCAL VOC 2012 dataset. Secondly, we present the results obtained on Cityscapes in subsection \ref{subsec: qualitative}, carrying out a qualitative and visual analysis of some of the most popular methods, showing some key examples where the performance of these methods can be observed.

\subsection{Quantitative results on PASCAL VOC 2012}
\label{subsec: quantitative}

In Table \ref{tab: results_pascal} we present the results obtained with the methods included in our experimentation. First, we show the fully supervised results that we obtain in each of the data partitions with the base model that later is used by the semi-supervised methods. Then, we present the results obtained with the different semi-supervised models that were presented in subsection \ref{sec:experiments} and represent all the categories of the defined taxonomy.

\begin{table}[!htp]\centering
\caption{ \small Semi-supervised SS and fully supervised baseline (DeepLabV3+) results on the PASCAL VOC 2012 dataset. Each column corresponds to a ratio of labeled/unlabeled images (the number on the left represents the number of labeled images used in each case). In each column the result obtained with the best performing method is highlighted. (Metric: mean IoU). }\label{tab: results_pascal}
\begin{tabular}{|l|c|c|c|c|c|}\hline
\textbf{Method} &\textbf{1/100 }(106) &\textbf{1/50 }(212) &\textbf{1/20} (529) &\textbf{1/8 }(1323) \\\hline \hline
DeepLabV3+  &48.75 &57.44 &66.17 &70.23 \\ 
Mean Teacher  &44.92 &58.53 &67.80 &71.55 \\
ClassMIX &56.45 &67.61 &70.78 &71.94 \\
CPS &47.70 &56.65 &69.59 &74.67 \\
CCT &37.05 &51.99 &62.26 &68.57 \\
s4GAN  &50.36 &62.31 &65.31 &71.26 \\
ST &55.20 &64.75 &71.37 &\textbf{74.88} \\
DMT &\textbf{58.94} &\textbf{70.05} &\textbf{72.29} &74.37 \\
ReCo &56.21 &63.20 &68.16 &72.50 \\
CAC &49.70 &64.30 &70.59 &74.59 \\

\hline
\end{tabular}
\end{table}


The first aspect to evaluate is the difference in performance between supervised and semi-sueprvised approaches. It is evident that semi-supervised approaches must show some improvement with respect to the supervised model that justifies the increase in complexity necessary to process the unlabeled data and extract knowledge from them. However, this requirement is not always fulfilled, and sometimes, in certain scenarios, the inclusion of unlabeled data in the training process can even harm the performance of the fully supervised model. The case in which this happens in the most extreme way is the CCT method, which obtains considerably worse results than the supervised model in all partitions. This behavior is accentuated to a greater extent as the amount of labeled data is reduced. Other methods, such as Mean Teacher, although they do not obtain as notable a deterioration as CCT, also present difficulties in extracting knowledge from unlabeled data, obtaining a gain between 1-2\% in all partitions, except in the partition with the least number of label data, in which it obtains worse results than the supervised model.

The next method in performance terms would be the s4GAN adversary method. This method obtains performance improvements in almost all the partitions with respect to the supervised model, these improvements varying from 1-5\%, except in the 1/20 partition, which does not manage to improve. Although it is true that a 5\% improvement could be a desirable improvement in many scenarios, this improvement does not occur in all partitions, presenting some instability in the results depending on the number of labeled data used. In addition, this method suffers from an increase in complexity compared to other simpler methods to carry out adversary training, which is hardly justifiable in view of the results obtained.

Other methods show variable results among the different labeling ratios. Some of the best performing methods when we have a very small set of labeled images are the ClassMix and ReCo methods, which obtain the second and third best results in the 1/100 partition, respectively. However, as we increase the number of labeled images, the margin of benefit that this method presents with respect to the supervised baseline is not so wide and there are many other methods that outperform it. Conversely, the CPS and CAC methods are two of the best performers in scenarios where we have many labeled images, obtaining the second and third best results in the 1/8 partition, respectively, and as with the previous methods, their performance suffers as the size of the labeled partition is reduced, even obtaining worse performance than the supervised baseline in the case of CPS. We can consider these methods as particularly useful in this specific scenario, but not as methods that obtain good overall performance.

Finally, methods based on pseudo-labeling have been shown to be the best performing ones. First, the ST method based on a simple self-training, has obtained the best result in the partition with the highest number of labeled images (1/8) in addition to obtaining competitive results in the rest of the partitions. On the other hand, the DMT method, based on mutual training, obtained the best results in all the partitions, except in the 1/8 partition, which obtained a result less than 1\% lower than the best model.

Therefore we conclude that the family of methods that provide the best performance are the pseudo-labeling based methods, and specifically the DMT method, which, according to this experimental evaluation, can be considered the current state-of-the-art in semi-supervised SS.

\subsection{Qualitative results on Cityscapes}
\label{subsec: qualitative}

In this subsection we carry out a qualitative and visual analysis of the results obtained on the Cityscapes dataset with some of the most popular state-of-the-art methods. The methods used in this analysis are DMT, ClassMix and s4GAN. In the following we visually show the segmentation maps predicted with each of these methods on some representative examples of the Cityscapes dataset, comparing them with the ground truth. In addition, in order to clearly and quickly identify the areas where these methods fail, we generate error masks in which we highlight in black color those areas where the model has predicted an incorrect label. White areas correspond to unlabeled zones in the dataset that are not taken into account in the learning process.

\definecolor{road}{RGB}{128, 64, 128}
\definecolor{sidewalk}{RGB}{244, 35, 232}
\definecolor{building}{RGB}{70, 70, 70}
\definecolor{wall}{RGB}{102, 102, 156}
\definecolor{fence}{RGB}{190, 153, 153}
\definecolor{pole}{RGB}{153, 153, 153}
\definecolor{traffic light}{RGB}{250, 170, 30}
\definecolor{traffic sign}{RGB}{220, 220, 0}
\definecolor{vegetation}{RGB}{107, 142, 35}
\definecolor{terrain}{RGB}{152, 251, 152}
\definecolor{sky}{RGB}{0, 130, 180}
\definecolor{person}{RGB}{220, 20, 60}
\definecolor{rider}{RGB}{255, 0, 0}
\definecolor{car}{RGB}{0, 0, 142}
\definecolor{truck}{RGB}{0, 0, 70}
\definecolor{bus}{RGB}{0, 60, 100}
\definecolor{train}{RGB}{0, 80, 100}
\definecolor{motorcycle}{RGB}{0, 0, 230}
\definecolor{bicycle}{RGB}{119, 11, 32}


\begin{figure}
\begin{subfigure}[h]{0.24\textwidth}
    \includegraphics[width=.99\textwidth]{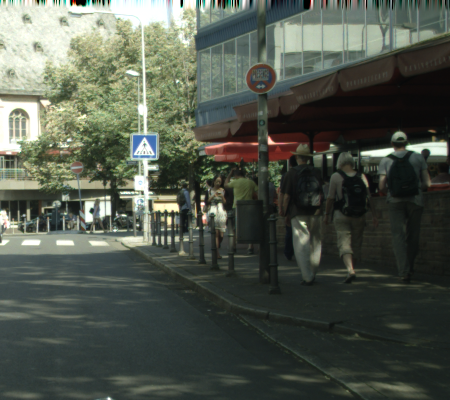}\caption{\scriptsize Original image}\hfill
\end{subfigure}
\begin{subfigure}[h]{0.24\textwidth}
    \includegraphics[width=.99\textwidth]{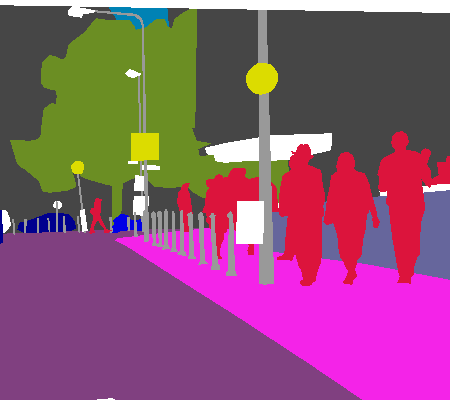}\caption{\scriptsize Ground truth}\hfill
\end{subfigure}

\begin{subfigure}[h]{0.157\textwidth}
    \includegraphics[width=.99\textwidth]{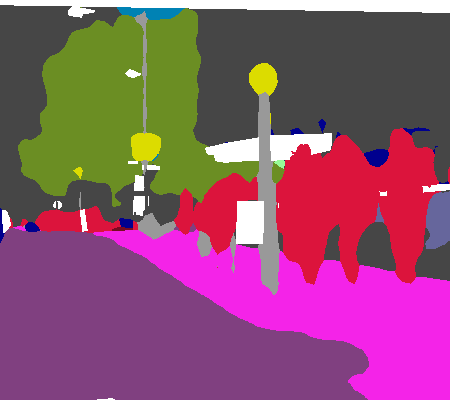}\caption{\scriptsize DMT prediction}\hfill
\end{subfigure}
\begin{subfigure}[h]{0.157\textwidth}
    \includegraphics[width=.99\textwidth]{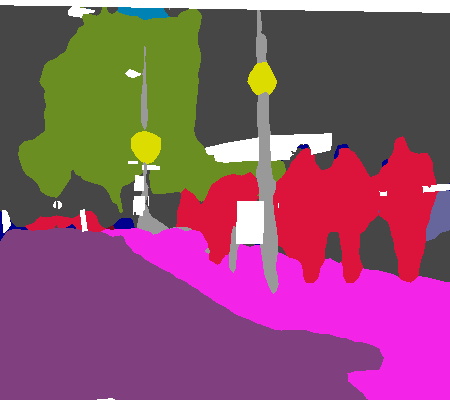}\caption{\scriptsize s4GAN prediction }\hfill
\end{subfigure}
\begin{subfigure}[h]{0.157\textwidth}
    \includegraphics[width=.99\textwidth]{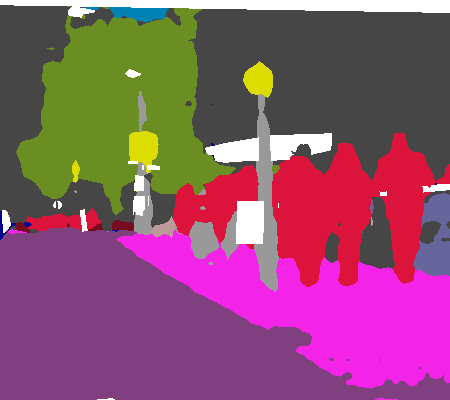}\caption{\scriptsize ClassMix prediction}\hfill
\end{subfigure}

\begin{subfigure}[h]{0.157\textwidth}
    \includegraphics[width=.99\textwidth]{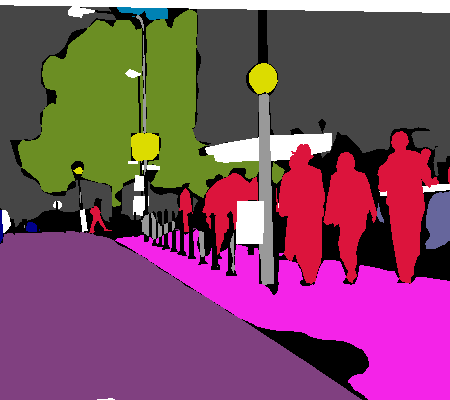}\caption{\scriptsize DMT error mask}\hfill
\end{subfigure}
\begin{subfigure}[h]{0.157\textwidth}
    \includegraphics[width=.99\textwidth]{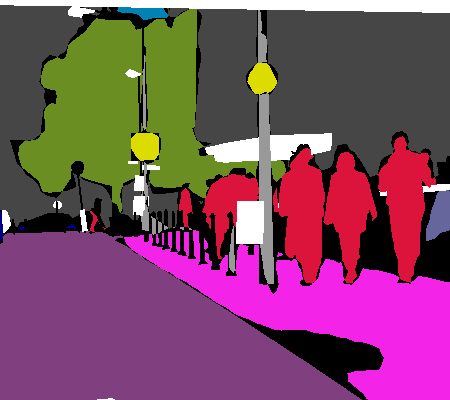}\caption{\scriptsize s4GAN error mask}\hfill
\end{subfigure}
\begin{subfigure}[h]{0.157\textwidth}
    \includegraphics[width=.99\textwidth]{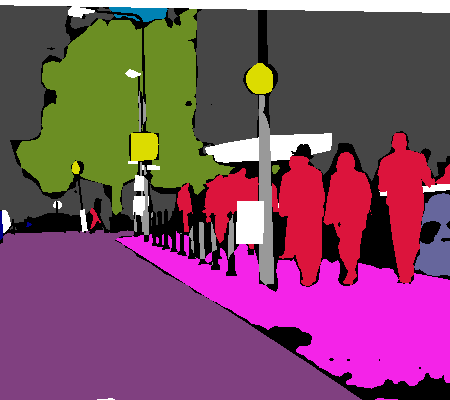}\caption{\scriptsize ClassMix error mask}\hfill
\end{subfigure}

    \caption{ \small Qualitative results obtained with DMT, s4GAN and ClassMix methods in an example of Cityscapes with main representation of the classes person, road, sidewalk, vegetation and building. Black color represents prediction errors.}\label{fig: visual1}
\end{figure}

In the first visual example shown in Figure \ref{fig: visual1} we can see a good and similar performance of the methods used in this qualitative analysis in the classes that predominate in the image, such as the road (\tikz\draw[road,fill=road] (0,0) circle (.7ex);), sidewalk (\tikz\draw[sidewalk,fill=sidewalk] (0,0) circle (.7ex);), building (\tikz\draw[building,fill=building] (0,0) circle (.7ex);) and vegetation (\tikz\draw[vegetation,fill=vegetation] (0,0) circle (.7ex);) classes. We only see an area in the lower right corner where the three models have a clear confusion between these predominant classes, specifically between the road and sidewalk classes, as we can see in the error masks, due to an irregularity in the sidewalk. Another largely represented class in this image is the person (\tikz\draw[person,fill=person] (0,0) circle (.7ex);) class. Although all models detect the presence of people, they have more difficulties when it comes to exactly defining the area belonging to each person. Unlike the previously named classes that usually appear in the image in a single large area, being easy to predict for the models, classes such as person, which present greater fragmentation by appearing in different and smaller areas of the image that do not have to be adjacent, suppose a greater difficulty for the models, obtaining less exact predictions and predicting incorrect classes in the gaps between different instances of the person class.

\begin{figure}
\begin{subfigure}[h]{0.24\textwidth}
    \includegraphics[width=.99\textwidth]{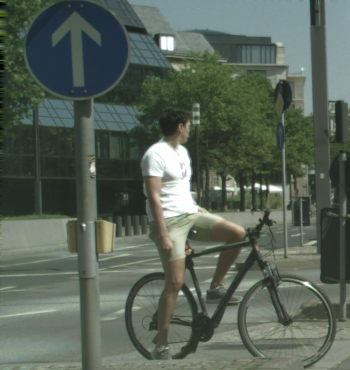}\caption{\scriptsize Original image}\hfill
\end{subfigure}
\begin{subfigure}[h]{0.24\textwidth}
    \includegraphics[width=.99\textwidth]{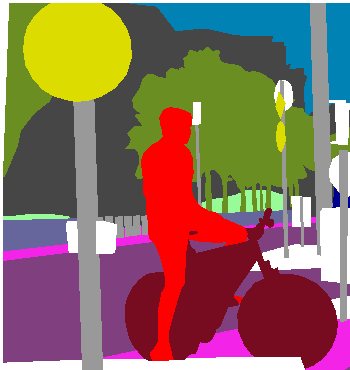}\caption{\scriptsize Ground truth}\hfill
\end{subfigure}

\begin{subfigure}[h]{0.157\textwidth}
    \includegraphics[width=.99\textwidth]{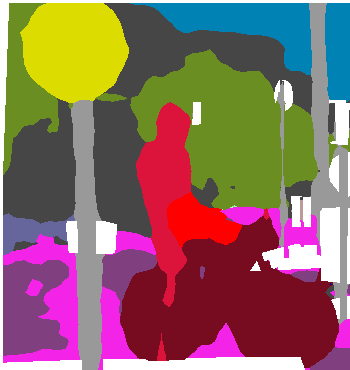}\caption{\scriptsize DMT prediction}\hfill
\end{subfigure}
\begin{subfigure}[h]{0.157\textwidth}
    \includegraphics[width=.99\textwidth]{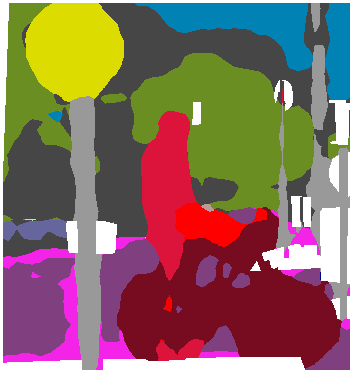}\caption{\scriptsize s4GAN prediction}\hfill
\end{subfigure}
\begin{subfigure}[h]{0.157\textwidth}
    \includegraphics[width=.99\textwidth]{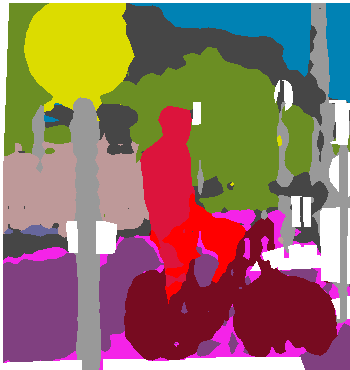}\caption{\scriptsize ClassMix prediction}\hfill
\end{subfigure}


\begin{subfigure}[h]{0.157\textwidth}
    \includegraphics[width=.99\textwidth]{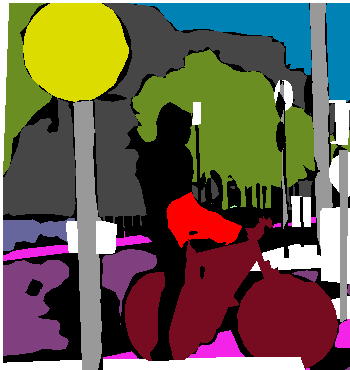}\caption{\scriptsize DMT error mask}\hfill
\end{subfigure}
\begin{subfigure}[h]{0.157\textwidth}
    \includegraphics[width=.99\textwidth]{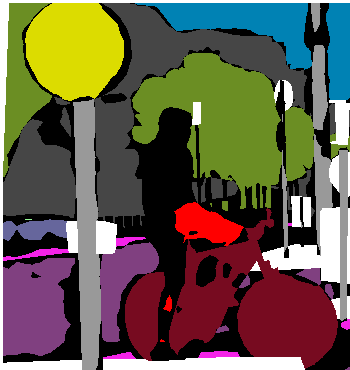}\caption{\scriptsize s4GAN error mask}\hfill
\end{subfigure}
\begin{subfigure}[h]{0.157\textwidth}
    \includegraphics[width=.99\textwidth]{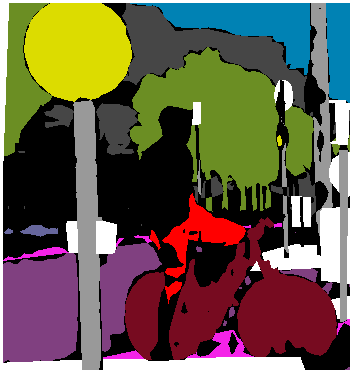}\caption{\scriptsize ClassMix error mask}\hfill
\end{subfigure}

    \caption{ \small Qualitative results obtained with DMT, s4GAN and ClassMix methods in an example of Cityscapes with main representation of the classes bicycle, rider, pole and traffic sign. Black color represents prediction errors.}\label{fig: visual3}
\end{figure}
 In a second example shown in Figure \ref{fig: visual3}, a generalized poor performance over the rider (\tikz\draw[rider,fill=rider] (0,0) circle (.7ex);) class can be observed. The rider class itself presents very few differences with respect to the person (\tikz\draw[person,fill=person] (0,0) circle (.7ex);) class. The way in which a model could differentiate a person from a rider would be to look at whether the rider is on a bicycle (\tikz\draw[bicycle,fill=bicycle] (0,0) circle (.7ex);) or motorcycle (\tikz\draw[motorcycle,fill=motorcycle] (0,0) circle (.7ex);) in the image. However, the results obtained seem to indicate that the models have problems when trying to learn these semantic relationships between classes or contextual information, confusing the instance of the rider class with the person class nearly in its totality. Only some of the parts of the person closest to the bike are segmented with the correct rider class. This indicates that the difficulty of learning semantic relations between objects is even greater as the distance between them increases. Additionally, in this example we can observe a good generalized performance in the classes traffic sign (\tikz\draw[traffic sign,fill=traffic sign] (0,0) circle (.7ex);), pole (\tikz\draw[pole,fill=pole] (0,0) circle (.7ex);), vegetation (\tikz\draw[vegetation,fill=vegetation] (0,0) circle (.7ex);) and sky (\tikz\draw[sky,fill=sky] (0,0) circle (.7ex);).







\section{Challenges and future trends}
\label{sec:challenges}


This section presents some of the main challenges related to the semi-supervised SS problem, as well as some of the most promising future research lines. 

\begin{itemize}
    
    \item \textbf{Evaluation standards.} Different studies we found in the semi-supervised SS literature do not present a homogeneous experimental framework (i.e. use of different datasets, different data partitions, different implementations or versions of the base model, etc.). The proposal of a standard and realistic experimental and evaluation framework that all researchers can adopt would be a key point in the development of this field of research.
    

    \item \textbf{Families of methods with improvement potential.} We highlight two categories which may have greater potential in future research. First, we highlight the pseudo-labeling methods, specifically the subcategory of mutual-training, which has obtained the best results in our experimental analysis. However, only two semi-supervised segmentation proposals exist in this subcategory, so we consider it to have great margin for improvement and development. Moreover, we also consider hybrid methods as an interesting category for future research, due to their novelty and possibilities of different combinations.
    
    \item \textbf{Diversity in base models.} Many of the methods studied employ more than one base model and the diversity of these models can be a key aspect to obtain a good final model. However, these methods are usually limited to choosing the state-of-the-art supervised segmentation model (i.e. DeepLabV3+ \cite{Chen2018EncoderDecoderWA} at present) obtaining a set of models poor in diversity, and no proposal attempts to go deeper into this decision. A possible future line of research could focus on the study of the implication of inter-model diversity on the final result of semi-supervised segmentation methods.
    
    \item \textbf{Evaluation on more realistic scenarios.} We have observed that some of the most widely used datasets in both the supervised and semi-supervised segmentation problem are object-centered image datasets (e.g., PASCAL VOC 2012). This type of images represent a very controlled scenario, which we are difficult to find in real-world problems. Models designed to obtain good results in this type of datasets may not be useful in real applications. New emerging datasets (e.g., Cityscapes) present less controlled images and more semantic dependencies between classes (a clear example of this type of semantic relationships can be seen in Figure \ref{fig: visual3}, between the rider and the bicycle). These types of datasets need new methods capable of dealing with less controlled images and modeling semantic dependencies between classes.
    
    \item \textbf{New trend: transformers.} Transformers \cite{Dosovitskiy2021AnII} are a specific type of network architecture, originally proposed for natural language processing problems, with a different philosophy than CNN. Recently, these models have started to be applied in CV problems, and specifically in SS. These models can learn semantic relationships between classes, even between those that appear far from each other within an image. This is desirable in real situations where such relationships are abundant. Despite transformers have recently started to be applied in supervised SS with promising results, only a few proposals have attempted to introduce them in the SSL scenario. As a consequence, the application of this new family of approaches in semi-supervised SS can be considered as one of the most promising future research lines.
\end{itemize}

\section{Conclusions}
\label{sec:conclusion}

This paper seeks to structure the knowledge generated in recent years, as well as to pose challenges and future research trends, around the rise of semi-supervised segmentation methods.

One of the main contributions of this paper is the proposal of a taxonomy, which classifies all previous works (a total of 43 recently published methods related to this field) into five categories: adversarial methods, consistency regularization, pseudo-labeling, constrastive learning and hybrid methods. In this manner, we provide the reader with a quick and precise way to know the state of the art in this field, as well as a detailed description of each of the existing methods.

The analysis of the state of the art and the defined taxonomy is complemented  with an experimental study that compares all taxonomic categories under homogeneous experimental conditions (employing the two most common datasets in the field: PASCAL VOC 2012 and Cityscapes). This allows the reader to have an intuition about the performance of each of them. This experimentation is composed of 10 methods, and we conclude as the method belonging to the mutual-training category (i.e. DMT) as the one that offers the best performance.

Finally, we reflect on the current challenges of semi-supervised segmentation and potential future lines of research, highlighting the need for standardization of the experimental and evaluation framework, the convenience of using realistic benchmarks where images are not controlled and are rich in semantic dependencies between classes, and the potential application in a semi-supervised scenario of a novel technology recently applied in CV, vision transformers.

\section*{Acknowledgments}
This work has been partially supported by the Contract UGR-AM OTRI-4260b. This work was also supported by project PID2020-119478 GB-I00 granted by Ministerio de Ciencia, Innovación y Universidades, and projects P18-FR-4961 by Proyectos I+D+i Junta de Andalucia 2018.
This work was also supported by the Spanish Ministry of Science and
Innovation, the Andalusian Government, and European Regional Development
Funds (ERDF) under grants CONFIA (PID2021-122916NB-I00) and FORAGE
(B-TIC-456-UGR20).
The hardware used in this work is supported by the projects with reference EQC2018-005084-P, granted by the Spain’s Ministry of Science and Innovation and European Regional Development Fund (ERDF) and  the project with reference SOMM17/6110/ UGR, granted by the Andalusian ‘’Consejería de Conocimiento, Investigación y Universidades’’ and European Regional Development Funds (ERDF).





\small
\bibliographystyle{IEEEtran}
\bibliography{main.bib}

\begin{IEEEbiography}[{\includegraphics[width=1in,height=1.25in,clip,keepaspectratio]{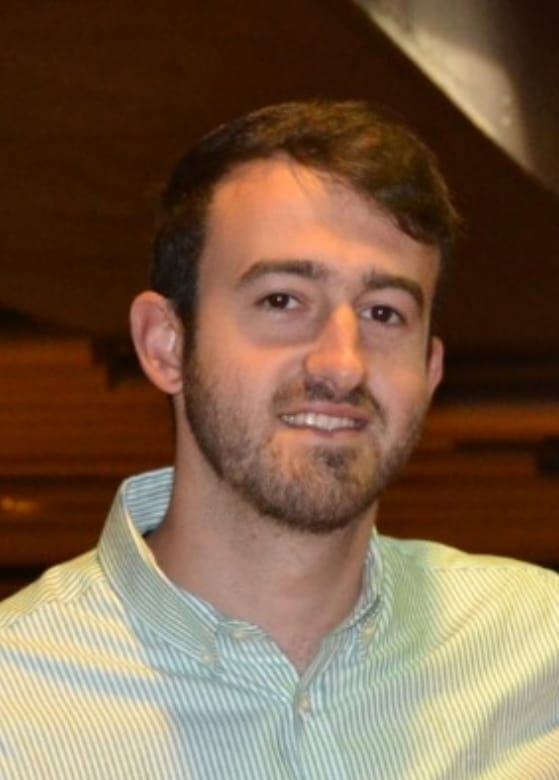}}]{Adrián Peláez}
received the BSc degree in computer engineering and MSc degree in data science and computer engineering from the University of Granada, Granada, Spain, in 2019 and 2020 respectively, and is currently working toward the PhD degree in the Department of Computer Science and Artificial Intelligence, University of Granada, Granada, Spain and in the Andalusian Research Institute on Data Science and Computational Intelligence (DaSCI). His main research interests are artificial intelligence, computer vision, semantic segmentation, and semi-supervised learning.
\end{IEEEbiography}
\vskip 0pt plus -1fil

\begin{IEEEbiography}[{\includegraphics[width=1in,height=1.25in,clip,keepaspectratio]{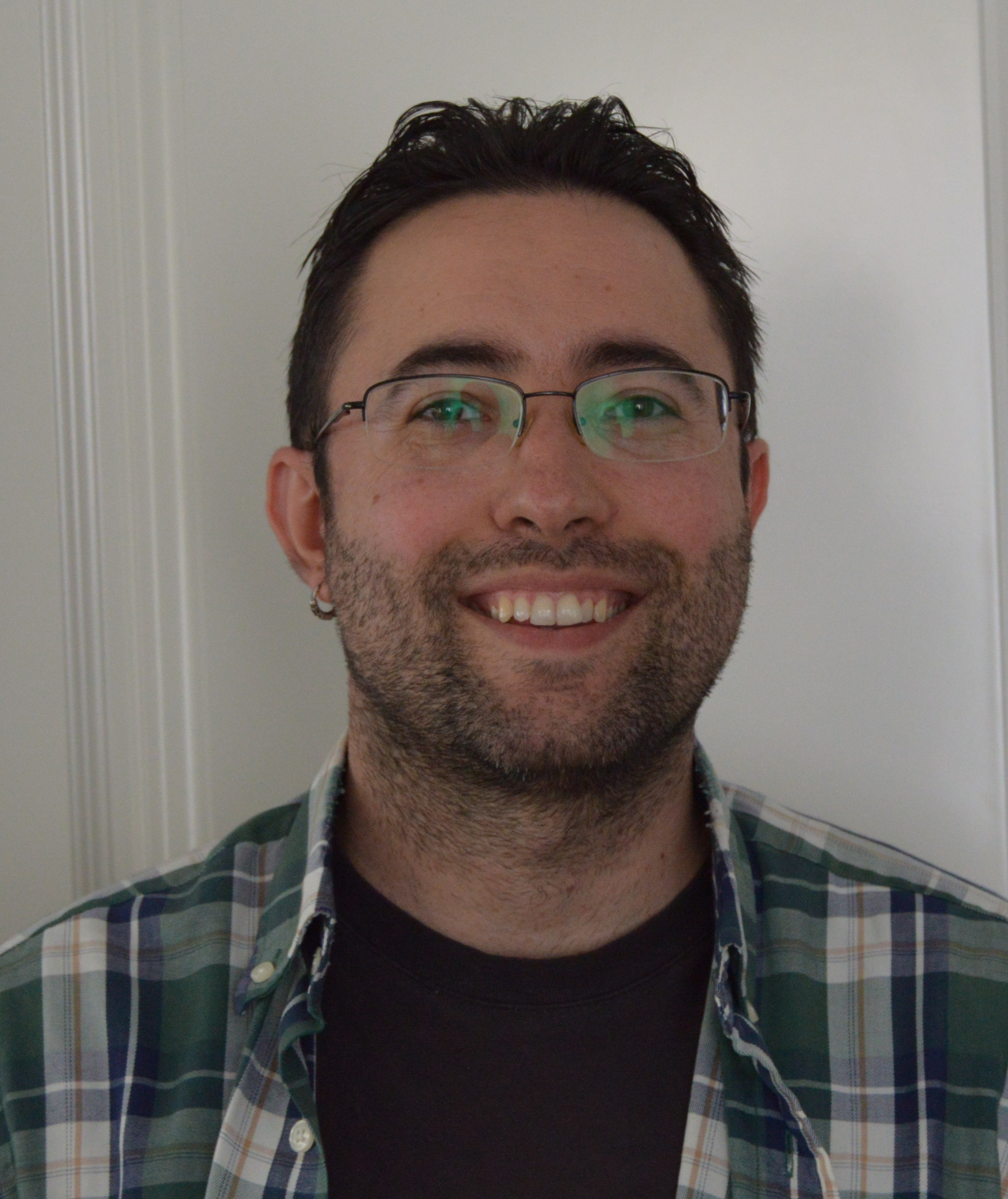}}]{Pablo Mesejo} is an Associate Professor at the Department of Computer
Science and Artificial Intelligence (DECSAI) of the University of Granada
(UGR, Spain). The main topic of his research is the analysis and design of
machine learning, computer vision and computational intelligence methods
able to solve image analysis problems, mainly related to the biomedical
domain. He is co-founding partner and chief AI officer of Panacea
Cooperative Research, vice-chair of the IEEE CIS Task Force on Evolutionary
Computer Vision and Image Processing (chair from 2018 to 2021), member of
the IEEE CIS Task Force on Evolutionary Deep Learning and Applications,
Associate Member of the American Academy of Forensic Sciences (AAFS,
Digital and Multimedia Sciences Section), and member of the Andalusian
Research Institute on Data Science and Computational Intelligence (DaSCI).
\end{IEEEbiography}
\vskip 0pt plus -1fil

\begin{IEEEbiography}[{\includegraphics[width=1in,height=1.25in,clip,keepaspectratio]{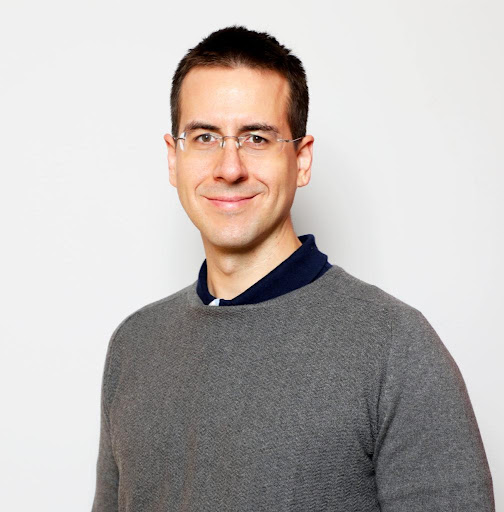}}]{Julián Luengo} received the M.S. degree in computer science and the Ph.D. from the University of Granada, Granada, Spain, in 2006 and 2011 respectively. He currently acts as an Associate Professor in the Department of Computer Science and Artificial Intelligence at the University of Granada, Spain. He has published more than 45 JCR papers and has been awarded as a Highly Cited Researcher in 2018.
His research interests include machine learning and data mining, data preparation in knowledge discovery and data mining, missing values, noisy data, data complexity and fuzzy systems. Recently, he has researched in the topic of Deep Learning segmentation and anomaly detection due to the increasing industry focus on these topics.

\end{IEEEbiography}
\end{document}